\documentclass[10pt,twocolumn,letterpaper]{article}

\usepackage{cvpr}
\usepackage{times}
\usepackage{epsfig}
\usepackage{graphicx}
\usepackage{amsmath}
\usepackage{amssymb}
\usepackage{url}
\usepackage{subfig,caption}
\usepackage{algorithm}
\usepackage{algpseudocode}
\usepackage{pifont}
\usepackage{color}

\cvprfinalcopy 

\def\cvprPaperID{1270} 

\ifcvprfinal\fi
\begin{document}

\title{Human Pose Estimation with Iterative Error Feedback}

\author{
Jo\~{a}o ~Carreira\thanks{Now at Google DeepMind.} \\
\texttt{carreira@eecs.berkeley.edu} \\
\texttt{UC Berkeley} \\
\and
Pulkit ~Agrawal \\
\texttt{pulkitag@eecs.berkeley.edu} \\
\texttt{UC Berkeley} \\
\and
Katerina ~Fragkiadaki\thanks{Now at Google.} \\
\texttt{katef@eecs.berkeley.edu} \\
\texttt{UC Berkeley}
\and
Jitendra
Malik \\
\texttt{malik@eecs.berkeley.edu} \\
\texttt{UC Berkeley}
}

\newcommand{\fix}{\marginpar{FIX}}
\newcommand{\new}{\marginpar{NEW}}
\newcommand{\control}{\textit{predictor}}
\newcommand{\corrgen}{\textit{correction generator}}
\newcommand{\pComment}[1]{\textcolor{red} {Pulkit: #1}}

\maketitle

\begin{abstract}

Hierarchical feature extractors such as Convolutional Networks (ConvNets) have achieved impressive performance on a variety of classification tasks using purely feedforward processing. Feedforward architectures can learn rich representations of the input space but do not explicitly model dependencies in the output spaces, that are quite structured for tasks such as articulated human pose estimation or object segmentation. Here we propose a framework that expands the expressive power of hierarchical feature extractors to encompass both input and output spaces, by introducing top-down feedback. Instead of directly predicting the outputs in one go, we use a self-correcting model that progressively changes an initial solution by feeding back error predictions, in a process we call Iterative Error Feedback (IEF). 
IEF  shows excellent performance on the task of articulated pose estimation in the challenging MPII and LSP benchmarks,  matching the state-of-the-art without requiring ground truth scale annotation. 

\end{abstract}

\section{Introduction}
\label{sec:intro}

Feature extractors such as Convolutional Networks (ConvNets) \cite{lecun1998gradient} represent images using a multi-layered hierarchy of features 
 and are inspired by the structure and functionality of the visual pathway of the human brain \cite{fukushima1980neocognitron, agrawal2014pixels}.
Feature computation in these models is purely feedforward, however, unlike in the human visual system where feedback connections abound \cite{citeulike:8460806,ventral,LammeRoelfaema00}. Feedback can be used to modulate and specialize feature extraction in early layers in order to model temporal and spatial context (e.g. \textit{priming} \cite{tulving1990priming}), to leverage prior knowledge about shape for segmentation and 3D perception, or simply for guiding visual attention to image regions relevant for the task under consideration. 

Here we are interested in using feedback to build predictors that can naturally handle complex, structured output spaces. We will use as running example the task of 2D human pose estimation \cite{YangR_CVPR_2011,toshev2014deeppose,Tompson_2015_CVPR,pfister2015flowing}, where the goal is to infer the 2D locations of a set of keypoints such as wrists, ankles, etc, from a single RGB image. The space of 2D human poses is highly structured because of body part proportions, left-right symmetries, interpenetration constraints, joint limits (e.g. elbows do not bend back) and physical connectivity (e.g.  wrists are rigidly related to elbows), among others. Modeling this structure should make it easier to pinpoint the visible keypoints and make it possible to estimate the occluded ones.

Our main contribution is in providing a generic framework for modeling rich structure in both input and output spaces by learning  hierarchical feature extractors over their joint space. We achieve this by incorporating top-down feedback -- instead of trying to directly predict the target outputs, as in feedforward processing, we predict what is wrong with their current estimate and correct it iteratively.  We call our framework Iterative Error Feedback, or IEF.

\begin{figure*}
\centering
\includegraphics[width=0.9\linewidth]{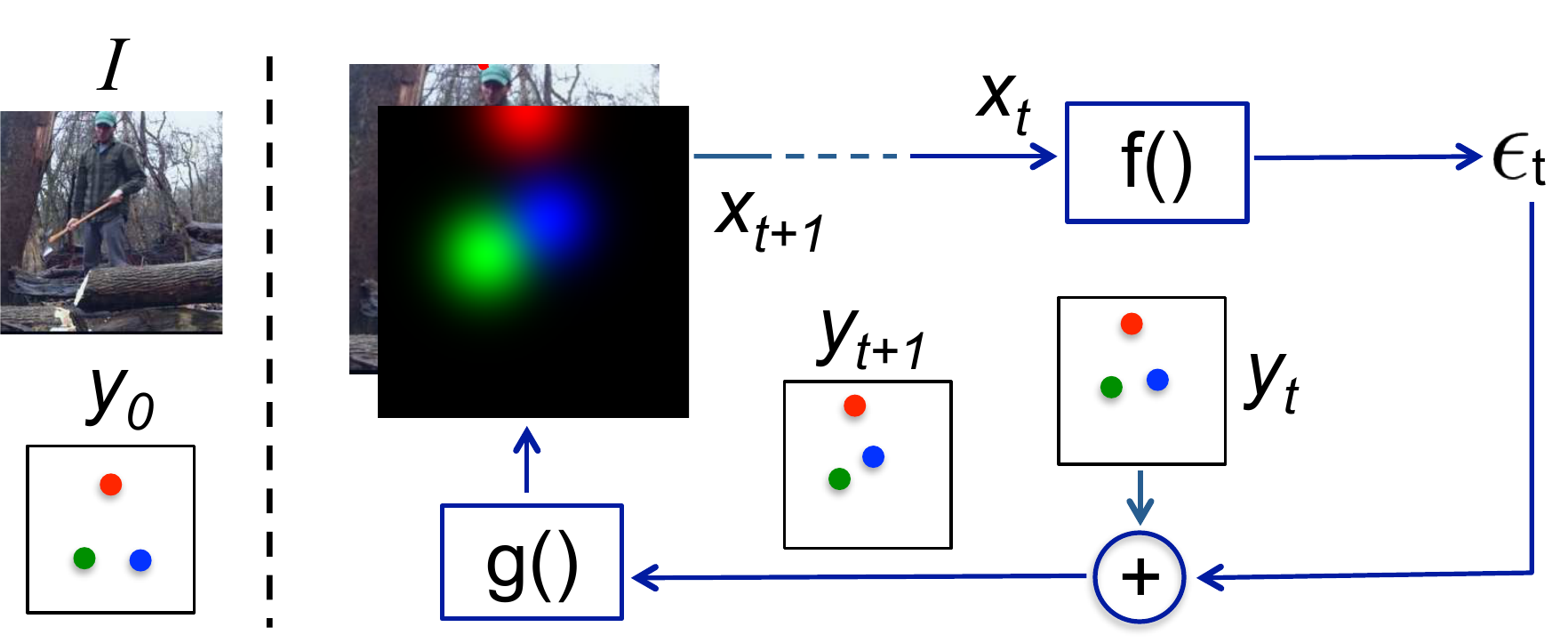} 
\caption{
An implementation of Iterative Error Feedback (IEF) for 2D human pose estimation. The left panel shows the input image $I$ and the initial guess of keypoints $y_0$, represented as a set of 2D points. For the sake of illustration we show only 3 out of 17 keypoints, corresponding to the right wrist (green), left wrist (blue) and top of head (red). Consider iteration $t$:  predictor $f$ receives the input $x_{t}$ -- image $I$ stacked with a ``rendering" of current keypoint positions $y_{t}$ -- and outputs a correction $\epsilon_t$. This correction is added to $y_{t}$, resulting in new keypoint position estimates $y_{t+1}$. The new keypoints are rendered by function $g$ and stacked with image $I$, resulting in $x_{t+1}$, and so on iteratively. Function $f$ was modeled here as a ConvNet. Function $g$ converts each 2D keypoint position into one Gaussian heatmap channel. For $3$ keypoints there are $3$ stacked heatmaps which are visualized as channels of a color image. 
In contrast to previous works, in our framework multi-layered hierarchical models such as ConvNets can learn rich models over the joint space of body configurations and images.}
\label{fig:illustrate}
\end{figure*}

In IEF, a feedforward model $f$ operates on the augmented input space created by concatenating (denoted by $\oplus$) the RGB image $I$ with a visual representation $g$ of the estimated output $y_t$ to predict a ``correction" ($\epsilon_t$) that brings $y_t$ closer to the ground truth output $y$. The correction signal $\epsilon_t$ is applied to the current output $y_t$ to generate $y_{t+1}$ and this is converted into a visual representation by $g$, that is stacked with the image to produce new inputs $x_{t+1} = I$ $\oplus$ $g(y_{t})$ for $f$, and so on iteratively. This procedure is initialized with a guess of the output ($y_0$) and is repeated until a predetermined termination criterion is met. The model is trained to produce bounded corrections at each iteration, e.g.  $||\epsilon_t||_2 < L$. The motivation for modifying $y_t$ by a bounded amount is that the space of $x_t$ is typically highly non-linear and hence local corrections should be easier to learn. The working of our model can be mathematically described by the following equations:
\begin{align}
\label{eq:model-f}
\epsilon _{t} = f(x_t) \\
\label{eq:correction}
y_{t+1} = y_t + \epsilon_t \\
\label{eq:model-g}
x_{t+1} = I \oplus g (y_{t+1}),
\end{align}

where functions $f$ and $g$ have additional learned parameters $\Theta _{f}$ and $ \Theta _{g}$, respectively. Although we have used the predicted error to additively modify $y_t$ in equation \ref{eq:correction}, in general $y_{t+1}$ can be a result of an arbitrary non-linear function that operates on $y_t, \epsilon_t$.

In the running example of human pose estimation, $y_t$ is vector of retinotopic positions of all keypoints that are individually mapped by $g$ into heatmaps (i.e. $K$ heatmaps for $K$ keypoints). The heatmaps are stacked together with the image and passed as input to $f$ (see figure \ref{fig:illustrate} for an overview). The ``rendering" function $g$ in this particular case is not learnt -- it is instead modelled as a 2D Gaussian having a fixed standard deviation and centered on the keypoint location.
Intuitively, these heatmaps encode the current belief in keypoint locations in the image plane and thus form a natural representation for learning features over the joint space of body configurations and the RGB image.

The dimensionality of inputs to $f$ is $H \times W \times(K+3)$, where $H$, $W$ represent the height and width of the image and $(K+3)$ correspond to $K$ keypoints and the $3$ color channels of the image.
We model $f$ with a ConvNet with parameters $\Theta _{f}$ (i.e. ConvNet weights). As the ConvNet takes  $I$ $\oplus$ $g(y_{t})$ as inputs, it has the ability to learn features over the joint input-output space.

\section{Learning}
In order to infer the ground truth output ($y$), our method iteratively refines the current output ($y_t$). At each iteration, $f$ predicts a correction ($\epsilon _t$) that locally improves the current output. 
Note that we train the model to predict bounded corrections, but we do not enforce any such constraints at test time. The parameters  ($\Theta_f, \Theta_g$) of functions $f$ and $g$ in our model, are learnt by optimizing equation \ref{eq:loss},

\begin{equation}
\label{eq:loss}
\min_{\Theta_f, \Theta_g} \sum_{t=1}^{T} h(\epsilon_t , e(y, y_t)) 
\end{equation}

where, $\epsilon_t$ and $e(y, y_t)$  are predicted and target bounded corrections, respectively. The function $h$  is a measure of distance, such as a quadratic loss. $T$ is the number of correction steps taken by the model. $T$ can either be chosen to be a constant or, more generally, be a function of $\epsilon_t$ (i.e. a termination condition). 

We optimize this cost function using stochastic gradient descent (SGD) with every correction step being an independent training example. We grow the training set progressively: we start by learning with the samples corresponding to the first step for $N$ epochs, then add the samples corresponding to the second step and train another $N$ epochs, and so on, such that early steps get optimized longer -- they get consolidated. 

As we only assume that the ground truth output ($y$) is provided at training time, it is unclear what the intermediate targets ($y_t$) should be. The simplest strategy, which we employ, is to predefine $y_t$ for every iteration using a set of fixed corrections $e(y, y_t)$ starting from $y_0$, obtaining ($y_0, y_1, ..y$). We call our overall learning procedure \textit{Fixed Path Consolidation (FPC)} which is formally described by algorithm \ref{fpc-learning}. 

\begin{figure*}[t]
\centering
\includegraphics[width = 0.95\textwidth]{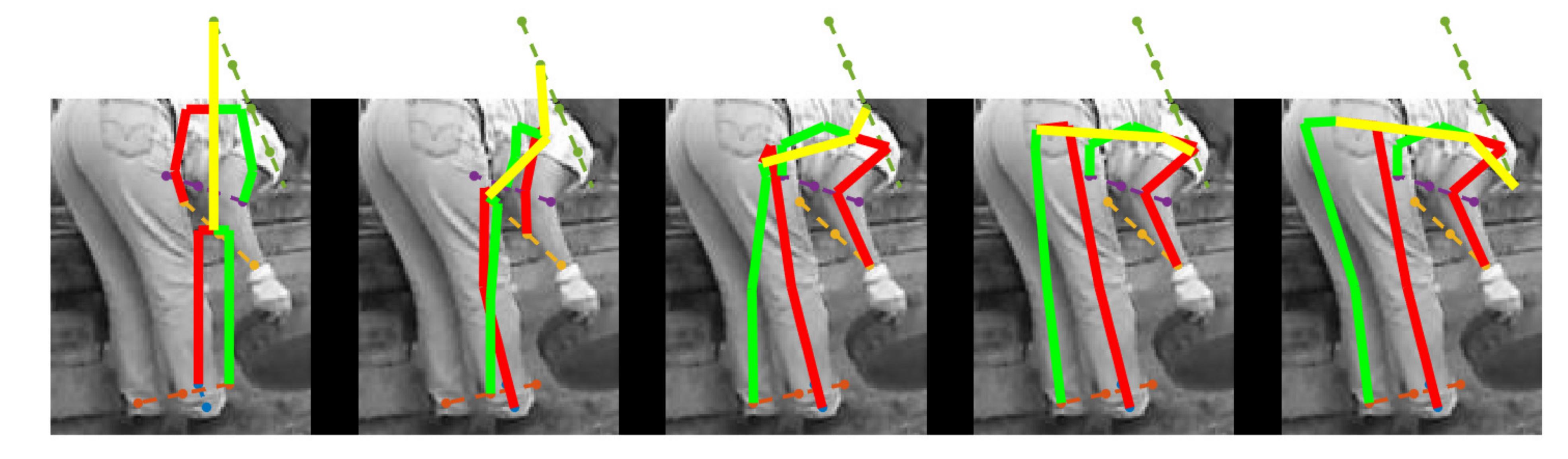}
\includegraphics[width = 0.95\textwidth]{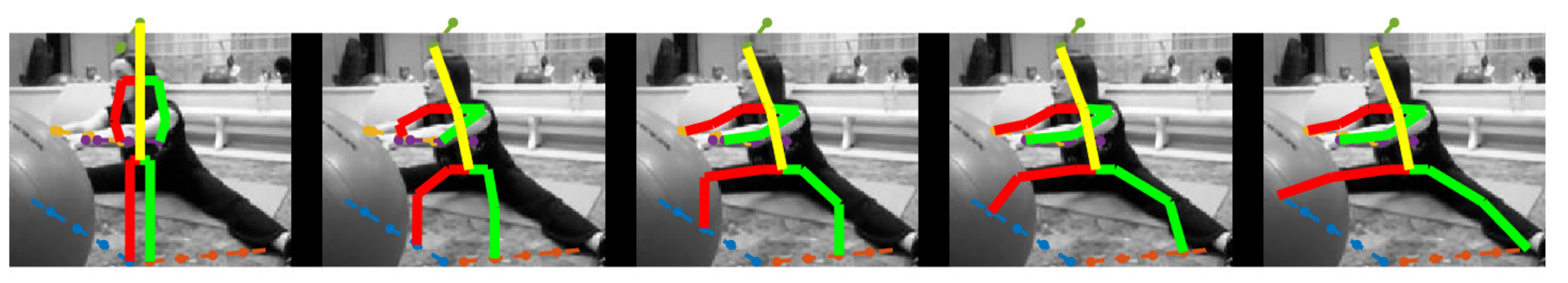}
  \caption{\label{fig:fpc} In our human pose estimation running example, the sequence of corrections $\epsilon_t$ moves keypoints along lines in the image, starting from an initial mean pose $y_0$ (left), all the way to the ground truth pose $y$ (right), here shown for two different images. This simplifies prediction at test time, because the desired corrections to each keypoint are constant for each image, up to the last one which is a scaled version. Feedback allows the model to detect when the  solution is close and to reduce "keypoint motion", as in a control system. Linear trajectories are shown for only a subset of the keypoints, to limit clutter.}
\end{figure*}

The target bounded corrections for every iteration are computed using a function $e(y, y_t)$, which can take different forms for different problems. If for instance the output is 1D, then $e(y, y_t) = max($sign$(y - y_t)\cdot \alpha$, $y - y_t)$ would imply that the target ``bounded" error will correct $y_t$ by a maximum amount of $\alpha$ in the direction of $y$.

\begin{algorithm}[t]
\caption{Learning Iterative Error Feedback with Fixed Path Consolidation}{} 
\label{fpc-learning}
\begin{algorithmic}[1]
\Procedure{FPC-Learn}{}
\State Initialize $y_0$
\State $E \leftarrow \{\}$
\For{$t \leftarrow 1 $ to $ (T_{steps})$ }
\For{all training examples $(I,y)$}
\State $\epsilon _{t} \leftarrow e(y,y_t)$
\EndFor
\State $E \leftarrow E \cup \epsilon _{t}$

\For{$j \leftarrow 1 $ to $ N$ }
\State Update  $\Theta _f$ and $\Theta _g$ with SGD, using loss $h$ and target corrections $E$
\EndFor
\EndFor
\EndProcedure
\end{algorithmic}
\end{algorithm}

\subsection{Learning Human Pose Estimation}
\label{sub:human-error}
Human pose was represented by a set of 2D keypoint locations $y: \{y^{k} \in \Re^{2},  k \in [1, K] \}$ where K is the number of keypoints and $y^k$ denotes the $k^{th}$ keypoint. The predicted location of keypoints at the $t^{th}$ iteration has been denoted by $y_t: \{y^{k}_t, k \in [1, K] \}$. The rendering of $y_t$ as heatmaps concatenated with the image was provided as inputs to a ConvNet (see section \ref{sec:intro} for details). The ConvNet was trained to predict a sequence of ``bounded" corrections for each keypoint ($\epsilon_{t}^k$) . The corrections were used to iteratively refine the keypoint locations.  

Let $u = y^k - y_{t}^{k}$ and the corresponding unit vector be $\hat{u} = \frac{u}{||u||_2}$. Then, the target ``bounded" correction for the $t^{th}$ iteration and $k^{th}$ keypoint  was calculated as:
\begin{equation}
\label{eq:error}
e(y^k, y_{t}^{k}) = \min (L, ||u||) \cdot \hat{u}
\end{equation}
where $L$ denotes the maximum displacement for each keypoint location. An interesting property of this function is that it is constant while a keypoint is far from the ground truth and varies only in scale when it is closer than $L$ to the ground truth. This simplifies the learning problem: given an image and a fixed initial pose, the model just needs to predict a constant direction in which to move keypoints, and to "slow down" motion in this direction when the keypoint becomes close to the ground truth. See  fig. \ref{fig:fpc} for an illustration.

The target corrections were calculated independently for each keypoint in each example and we used an $L_2$ regression loss to model $h$ in eq. \ref{eq:loss}.  We set $L$ to $20$ pixels in our experiments. We initialized $y_0$ as the median of ground truth 2D keypoint locations on training images and trained a model for  $T = 4$ steps, using $N = 3$ epochs for each new step. We found the fourth step to have little effect on accuracy and used $3$ steps in practice at test time. 

\vspace{4mm}
\noindent \textbf{ConvNet architecture.} We employed a standard ConvNet architecture pre-trained on Imagenet: the very deep googlenet \cite{Szegedy_2015_CVPR} \footnote{The VGG-16 network \cite{simonyan2014very} produced similar results, but required significantly more memory.}. We modified the filters in the first convolution layer (conv-1) to account for $17$ additional channels due to 17 keypoints. In our model, the conv-1 filters operated on 20 channel inputs. The weights of the first three conv-1 channels (i.e. the ones corresponding to the image) were initialized using the weights learnt by pre-training on Imagenet. The weights corresponding to the remaining 17 channels were randomly initialized with Gaussian noise of variance 0.1. We discarded the last layer of 1000 units that predicted the Imagenet classes and replaced it with a layer containing $32$ units, encoding the continuous $2D$ correction \footnote{Again, we do not bound explicitly the correction at test time, instead the network is taught to predict bounded corrections.} expressed in Cartesian coordinates (the 17th "keypoint" is the location of one point  anywhere inside a person, marking her,  and which is provided as input both during training and testing, see section \ref{sub:data}). We used a fixed  ConvNet input size of $224\times224$.

\section{Results}
We tested our method on the two most challenging benchmarks for 2D human pose estimation:
the MPII Human Pose dataset \cite{andriluka20142d}, which features significant scale variation, occlusion, and multiple people interacting, and Leeds Sports Pose dataset (LSP) \cite{Johnson10} which features complex poses of people in sports. 
For each person in every image, the goal is to predict the 2D locations of all its annotated keypoints.

\vspace{2mm}
\noindent \textbf{MPII -- Experimental Details}. 
\label{sub:data}
Human pose is represented as a set of 16 keypoints. An additional \textit{marking-point} in each person is available both for training and testing, located  somewhere inside each person's boundary. We  represent this point as an additional channel and stack it with the other 16 keypoint channels and the 3 RGB channels that we feed as input to a ConvNet. We used the same publicly available train/validation splits of  \cite{Tompson_2015_CVPR}. We evaluated the accuracy of our algorithm on the validation set using the standard PCKh metric \cite{andriluka20142d}, and also submitted results for evaluation on the test set once, to obtain the final score.

We cropped $9$ square boxes centered on the marking-point of each person, sampled uniformly over scale, from $1.4\times$ to $0.3\times$ of the smallest side of the image and resized them to $256\times256$ pixels. Padding was added as necessary for obtaining these dimensions and the amount of training data was further doubled by also mirroring the images. We used the ground truth height of each person at training time, which is provided on MPII, and select as training examples the $3$ boxes for each person having a side closest to $1.2\times$ the person height in pixels. We then trained googlenet models on random crops of $224\times224$ patches, using $6$ epochs of consolidation for each of $4$ steps. At test time, we predict which one of the $9$ boxes is closest to $1.2\times$ the height of the person in pixels, using a shallower model, the VGG-S ConvNet \cite{Chatfield14}, trained for that task using an $L_2$ regression loss. We then align our model to the center $224\times224$ patch of the selected window. The MatConvnet library \cite{vedaldi2014matconvnet} was employed for these experiments.

We train our models using keypoint positions for both visible and occluded keypoints, which MPII provides in many cases whenever they project on to the image (the exception are people truncated by the image border). We zero out the backpropagated gradients for missing keypoint annotations. Note that often keypoints lie outside the cropped image passed to the ConvNet, but this poses no issues to our formulation -- keypoints outside the image can be predicted and are still visible to the ConvNet as tails of rendered Gaussians.

\begin{figure}[h!]
  \centering
      \includegraphics[width=0.49\textwidth]{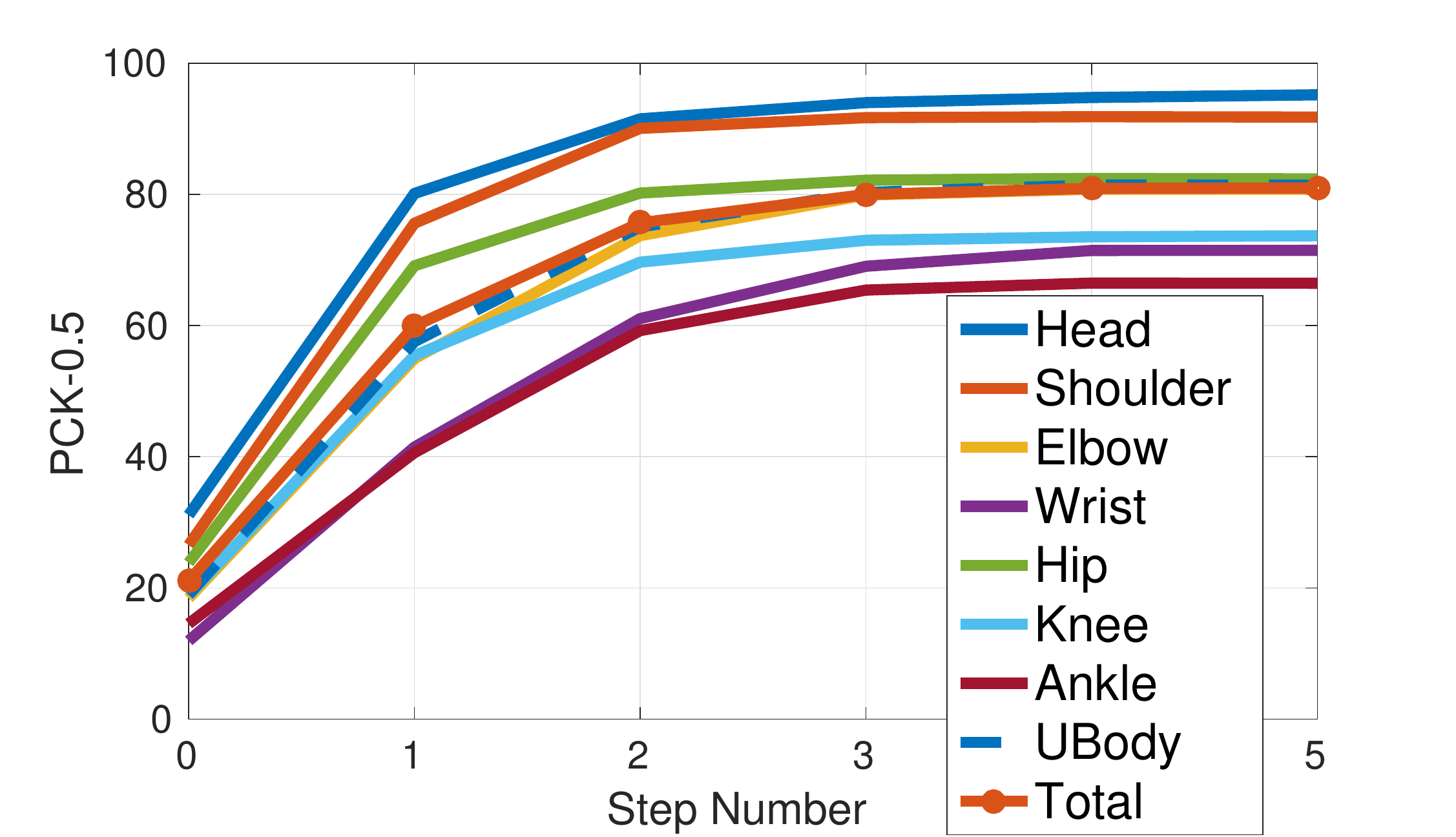}
  \caption{\label{ev}Evolution of PCKh at 0.5 overlap as function of correction step number on the MPII-human-pose validation set, using the finetuned googlenet network. The model aligns more accurately to parts like the head and shoulders, which is natural, because these parts are easier to discriminate from the background and have more consistent appearance than limbs.}
\end{figure}

\noindent \textbf{Comparison with State-of-the-Art.}  
The standard evaluation procedure in the MPII benchmark assumes ground truth scale information is known and images are normalized using this scale information. 
The current  state-of-the-art is the sliding-window approach of Tompson et al  \cite{Tompson_2015_CVPR} and IEF  roughly matches this performance, as shown in table \ref{numbers}.
In the more realistic setting of unknown scale information, the best previous  result so far is from Tompson et al.  \cite{Tompson_2015_CVPR} which was the first work to experiment with this setting and obtained $66.0$ PCKh. IEF significantly improves upon this number to $81.3$. Note however that the emphasis in Tompson et al's system was efficiency and they trained and tested their model using original image scales -- searching over a multiscale image pyramid or using our automatic rescaling procedure should presumably improve their performance. See the MPII website for more detailed results.

\begin{table*}
\footnotesize
\centering
\begin{tabular}{c|l*{7}{c}|c}
                  & Head & Shoulder & Elbow & Wrist & Hip  & Knee & Ankle & UBody & FBody\\
\hline
Yang \& Ramanan \cite{YangR_CVPR_2011} & 73.2 & 56.2 & 41.3 & 32.1 & 36.2 & 33.2 & 34.5 & 43.2 & 44.5 \\
Pischulin et al \cite{pishchulin2013poselet} & 74.2 & 49.0 & 40.8 & 34.1 & 36.5 & 34.4 & 35.1 & 41.3 & 44.0 \\ 
Tompson et al. \cite{Tompson_2015_CVPR}   & 96.1 & 91.9      & 83.9  & 77.8  & 80.9 & 72.3 & 64.8  & 84.5        & 82.0  \\
IEF    &    95.7 &  91.6  & 81.5  & 72.4  & 82.7 & 73.1 & 66.4 &  82.0  & 81.3 \\
\hline
Tompson et al. \cite{Tompson_2015_CVPR}   & 83.4 & 77.5     & 67.5  & 59.8  & 64.6 & 55.6 & 46.1  & 68.3        & 66.0  \\
IEF              &    95.5 &  91.6  & 81.5 & 72.4 & 82.7 & 73.1 & 66.9 & 81.9  & 81.3 \\
\end{tabular}
\caption{\label{numbers} MPII test set PCKh-0.5 results for Iterative Error Feedback (IEF) and previous approaches, when ground truth scale information at test time is provided (top) and in the more  automatic setting when it is not available (bottom).  UBody and FBody stand for upper body and full body, respectively.}
\end{table*}

\vspace{2mm}
\noindent \textbf{LSP -- Experimental Details}. \label{sub:data_lsp}  
In LSP, differently from MPII, images are usually tight around the person whose pose is being estimated, are resized so people have a fixed size, and have lower resolution. There is also no marking point on the torsos so we initialized the 17th keypoints used in MPII to the center of the image. The same set of keypoints is evaluated as in MPII and we trained a model using the same hyper-parameters on the extended LSP training set. We use the standard LSP evaluation code supplied with the MPII dataset and report person-centric PCP scores in table \ref{lsp_numbers}. Our results are competitive with the current state-of-the-art of Chen and Yuille \cite{chen2014articulated}. 

\begin{table*}
\footnotesize
\centering
\begin{tabular}{c|l*{5}{c}|c}
& Torso & Upper Leg & Lower Leg & Upper Arm & Forearm & Head  & Total \\
\hline
Pishchulin et al. \cite{pishchulin2013strong} & 88.9  & 64.0  & 58.1  & 45.5  & 35.1  & 85.1 & 58.0 \\
Tompson et al. \cite{NIPS2014_5573} & 90.3  & 70.4  & 61.1  & 63.0  & 51.2  & 83.7 & 66.6 \\
Fan et al. \cite{fan2015combining} & 95.4  & 77.7  & 69.8  & 62.8  & 49.1  & 86.6 & 70.1 \\
Chen and Yuille \cite{chen2014articulated} & 96.0  & 77.2  & 72.2  & 69.7  & 58.1  & 85.6 & 73.6 \\
IEF& 95.3  & 81.8  & 73.3  & 66.7  & 51.0  & 84.4 & 72.5 \\
\end{tabular}
\caption{\label{lsp_numbers} Person-centric PCP scores on the LSP dataset test set for IEF and previous approaches.}
\end{table*}

\section{Analyzing IEF}
\vspace{2mm}
In this section, we perform extensive ablation studies to validate four choices of the IEF model: 1)  proceeding iteratively instead of in a single shot, 2) predicting bounded corrections instead of directly predicting the target outputs, 3) curriculum learning of our bounded corrections, and 4) modeling the structure in the full output space (all body joints in this case)  over carrying out independent predictions for each label.

\begin{table*}
\footnotesize
\centering
\begin{tabular}{c|l*{7}{c}|c}
                  & Head & Shoulder & Elbow & Wrist & Hip  & Knee & Ankle & UBody & FBody\\
\hline
Iterative Error Feedback (IEF)  &    95.2 &  91.8  & 80.8  & 71.5  & 82.3  & 73.7  & 66.4 & 81.4  & 81.0 \\
Direct Prediction & 92.9 & 89.4 & 74.1 & 61.7 & 79.3 & 64.0 & 53.3 & 75.1 & 74.8 \\
Iterative Direct Prediction & 91.9  & 88.5 & 73.3  & 59.9 & 77.5  & 61.2  & 51.8 & 74.0  & 73.4  \\
\end{tabular}
\caption{\label{ablation} PCKh-0.5 results on the MPII validation set for models finetuned from googlenet using Iterative Error Feedback (IEF), direct regression to the keypoint locations (direct prediction), and a model that was trained to iteratively predict human pose by regressing to the ground truth keypoint locations (instead of bounded corrections) in each iteration, starting from the pose in the previous iteration. The results show that our proposed approach results in significantly better performance.}
\end{table*}

\vspace{2mm}
\noindent \textbf{Iterative v/s Direct Prediction.} For evaluating the importance of progressing towards solutions iteratively we trained models to directly predict corrections to the keypoint locations in a single shot (i.e. direct prediction). 
Table \ref{ablation} shows that IEF that additively regresses to keypoint locations achieves PCKh-0.5 of 81.0 as compared to PCKh of 74.8 achieved by directly regressing to the keypoints.

\vspace{2mm}
\noindent \textbf{Iterative Error Feedback v/s Iterative Direct Prediction.} Is iterative prediction of the error important or iterative prediction of the target label  directly (as in e.g.,  \cite{wolpert1992stacked,tu2008auto}) performs comparably?  
In order to answer this question we trained a model from the pretrained googlenet to iteratively predict the ground truth keypoint locations (as opposed to predicting bounded corrections). For comparing performance, we used the same number of iterations for this baseline model and IEF. Table \ref{ablation} shows that IEF achieves PCKh-0.5 of 81.0 as compared to PCKh of 73.4 by iterative direct prediction. This can be understood by the fact that the learning problem in IEF is much easier. In IEF, for a given image, the model is trained to predict constant corrections except for the last one which is a scaled version. In iterative direct prediction, because each new pose estimate ends up somewhere around the ground truth, the model must learn to adjust directions and magnitudes in all correction steps. 
\begin{figure}[h!]
  \centering
      \includegraphics[width=0.49\textwidth]{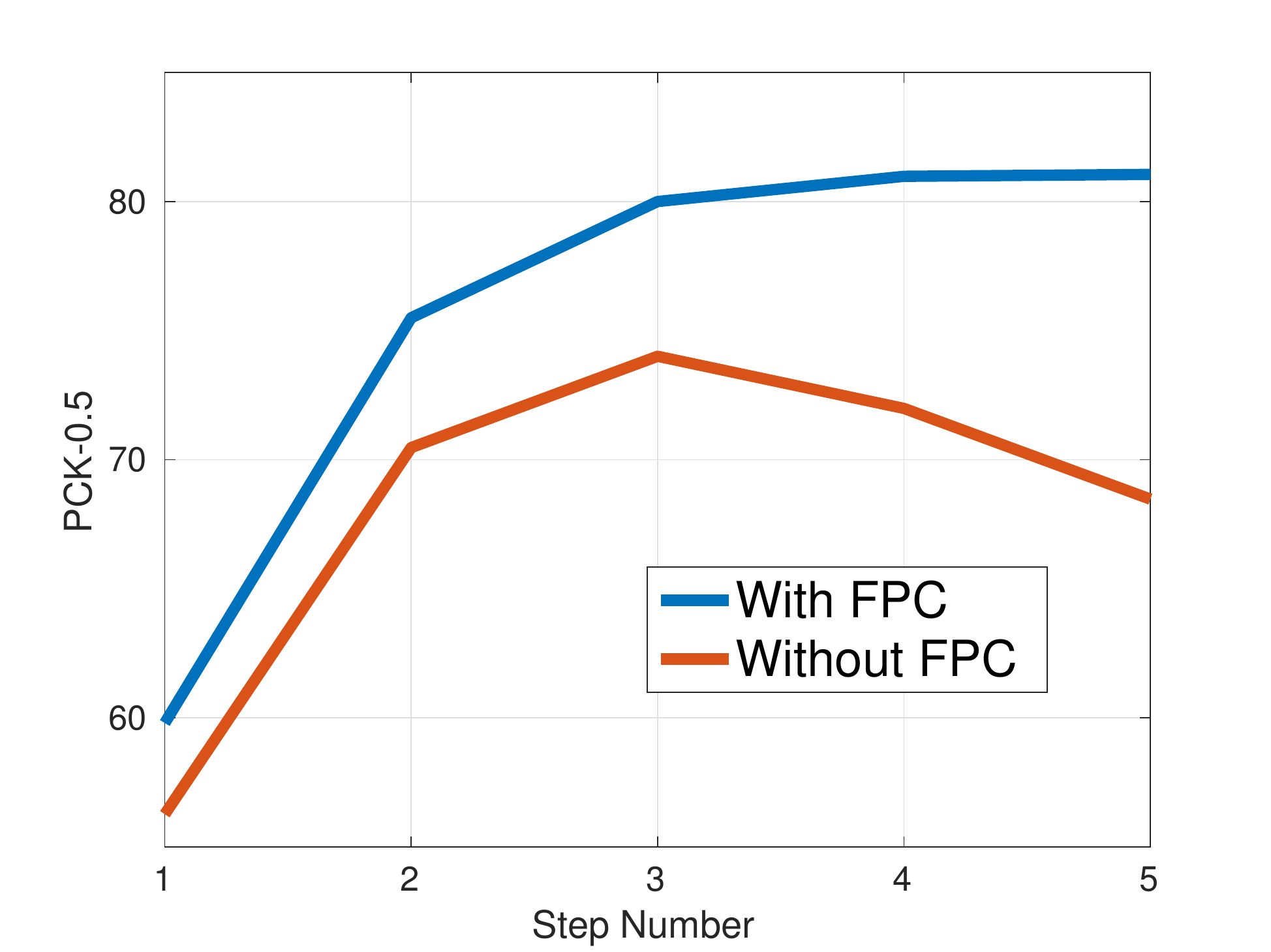}
  \caption{\label{training} Validation PCKh-0.5 scores for different number of correction steps taken, when finetuning a IEF model from a googlenet base model using stochastic gradient descent with either Fixed Path Consolidation (\textit{With FPC}), or directly over all training examples (\textit{Without FPC}), for the same amount of time. FPC leads to significantly more accurate results, leading to models that can perform more correction steps without drifting. It achieves this by consolidating the learning of earlier steps and progressively increasing the difficulty of the training set by adding additional correction steps.}
\end{figure}

\vspace{2mm}
\noindent \textbf{Importance of Fixed Path Consolidation (FPC).} The FPC method (see algorithm \ref{fpc-learning}) for training a IEF model  makes $N$ corrections is a curriculum learning strategy where in the $i^{th} (i \leq N)$ training stage the model is optimized for performing only the first $i$ corrections. Is this curriculum learning strategy necessary or can all the corrections be simultaneously trained? For addressing this question we trained an alternative model that trains for all corrections in all epochs. We trained IEF with and without FPC for the same number of SGD iterations and the performance of both these models is illustrated in figure \ref{training}. The figure shows that without FPC, the performance drops by almost 10 PCKh points on the validation set and that there is significant drift when performing several correction steps.

\vspace{2mm}
\noindent \textbf{Learning Structured Outputs.} One of the major merits of IEF is supposedly that it can jointly learn the structure in input images and target outputs. For human pose estimation, IEF models the space of outputs by augmenting the image with  additional input channels having gaussian renderings centered around estimated keypoint locations . If it is the case that IEF learns priors over the appropriate relative locations of the various keypoints,  then depriving the model of keypoints other than the one being predicted should decrease performance. 

In order to evaluate this hypothesis we trained three different IEF models and tested how well each predicted the location of the ``Left Knee" keypoint. The first model had only one input channel corresponding to the left knee, the second model had two channels corresponding to left knee and the left hip. The third model was trained using all keypoints in the standard IEF way. The performance of these three models is reported in table \ref{keypoint-ablation}. As a baseline, regression gets 64.6, whereas the IEF model with a single additional input channel for the left knee gets PCKh of 69.2 This shows that feeding back the current estimate of the left knee keypoint allows for more accurate localization by itself. Furthermore, the IEF model over both left knee and left hip  gets PCKh of 72.8. This suggests that the relationship between neighboring outputs has much of the information, but modeling all joints together with the image still wins, obtaining a PCKh of 73.8. 


\begin{table*}
\footnotesize
\centering
\begin{tabular}{c|c|c|c|c}
              & Direct Prediction of All Joints & IEF Left Knee & IEF Left Knee + Left Hip & IEF All Joints \\
\hline
Left Knee PCKh-0.5  & 64.6  &  69.2 & 72.8 & 73.8 \\
\end{tabular}
\caption{\label{keypoint-ablation} MPII validation  PCKh-0.5 results for left knee localization when using IEF and both training and predicting different subsets of joints. We also show the result obtained using a direct prediction variant similar to plain regression on all joints (having the mean pose Gaussian maps in the input). Modeling global body structure jointly with the image leads to best results by "IEF All Joints". Interestingly, feedback seems to add value by itself and IEF on the left knee, in isolation, significantly outperforms the direct prediction baseline.}
\end{table*}

\section{Related Work}

There is a rich literature on structured output learning \cite{tsochantaridis2004support,daume2009search} (e.g. see references in \cite{nowozin2011structured}) but it is a relatively modern topic in conjunction with feature learning, for computer vision  \cite{chen2014learning, jaderberg2014deep,Tompson_2015_CVPR,lecun1998gradient}.

Here we proposed a feedback-based framework for structured-output learning. Neuroscience models of the human brain suggest that feedforward connections act as information carriers while numerous feedback connections act as modulators or competitive inhibitors to aid feature grouping \cite{Gilbert2007677}, figure-ground segregation \cite{citeulike:506679} and object recognition \cite{wyatte:the}. In computer vision, feedback has been primarily used so far for learning selective attention \cite{NIPS2014_5542}; in \cite{NIPS2014_5542} attention is implemented by estimating a bounding box in an image for the algorithm to process next, while in \cite{stollenga2014deep} attention is formed by selecting some convolutional features over others (it does not have a spatial dimension).  

Stacked inference methods  \cite{ramakrishna2014pose,weiss2012structured,wolpert1992stacked,tu2008auto} are another related family of methods. Differently, some of these methods consider each output in isolation \cite{toshev2014deeppose}, all use different weights or learning models in each stage of inference \cite{Tompson_2015_CVPR} or they do not optimize for correcting their current estimates but rather attempt to predict the answer from scratch at each stage \cite{li2013fixed,tu2008auto}. 
In concurrent work, Oberweger et al \cite{oberweger2015training} proposed a feedback loop for hand pose estimation from kinect data that is closely related to our approach. The autocontext work of \cite{tu2008auto} is also related and iteratively computes label heatmaps by concatenating the image with the heatmaps previously predicted. IEF  is inspired by this work and we show how this iterative computation can be carried out effectively with deep Convnet architectures, and with bounded error corrections,  rather than  aiming for the answer from scratch at each iteration. 






Another line of work aims to inject class-specific spatial priors using coarse-to-fine processing, e.g. features arising from different layers of ConvNets were recently used for instance segmentation and keypoint prediction \cite{hariharan2014hypercolumns}. For pose inference, combining multiple scales \cite{Fan_2015_CVPR,Tompson_2015_CVPR} aids in capturing subtle long-range dependencies (e.g. distinguishing the left and right sides of the body which depend on whether a person is facing the camera).  The system in our human pose estimation example can be seen as closest to approaches employing ``pose-indexed features'' \cite{fleuret-geman-rr2007,DollarCVPR10pose,ionescu2014iterated}, but leveraging hierarchical feature learning. Graphical models can also encode dependencies between outputs and are still popular in many applications, including human pose estimation \cite{chen2014articulated}.




Classic spatial alignment and warping computer vision models, such as snakes, \cite{kass1988snakes} and Active Appearance Models (AAMs) \cite{cootes1995active} have similar goals as the proposed IEF, but are not learned end-to-end -- or learned at all -- employ linear shape models and hand designed features and require slower gradient computation which often takes many iterations before convergence. They can get stuck in poor local minimas even for constrained variation (AAMs and small out-of-plane face rotations). IEF, on the other hand, is able to minimize over rich articulated human 3D pose variation, starting from a mean shape.  Although extensions that use learning to drive the optimization have been proposed \cite{xiong2013supervised}, typically these methods still require manually defined energy functions to measure goodness of fit.

\begin{figure*}
\centering
\renewcommand{\arraystretch}{1}
\begin{tabular}{@{}c@{\hspace{1pt}}c@{\hspace{1pt}}c@{\hspace{1pt}}c@{\hspace{1pt}}c}
{\small Mean shape} & {\small Step 1} & {\small Step 2} & {\small Step 4} & {\small Ground Truth} \\
\includegraphics[height = 0.1\textheight, width = 0.2\textwidth, keepaspectratio = true]{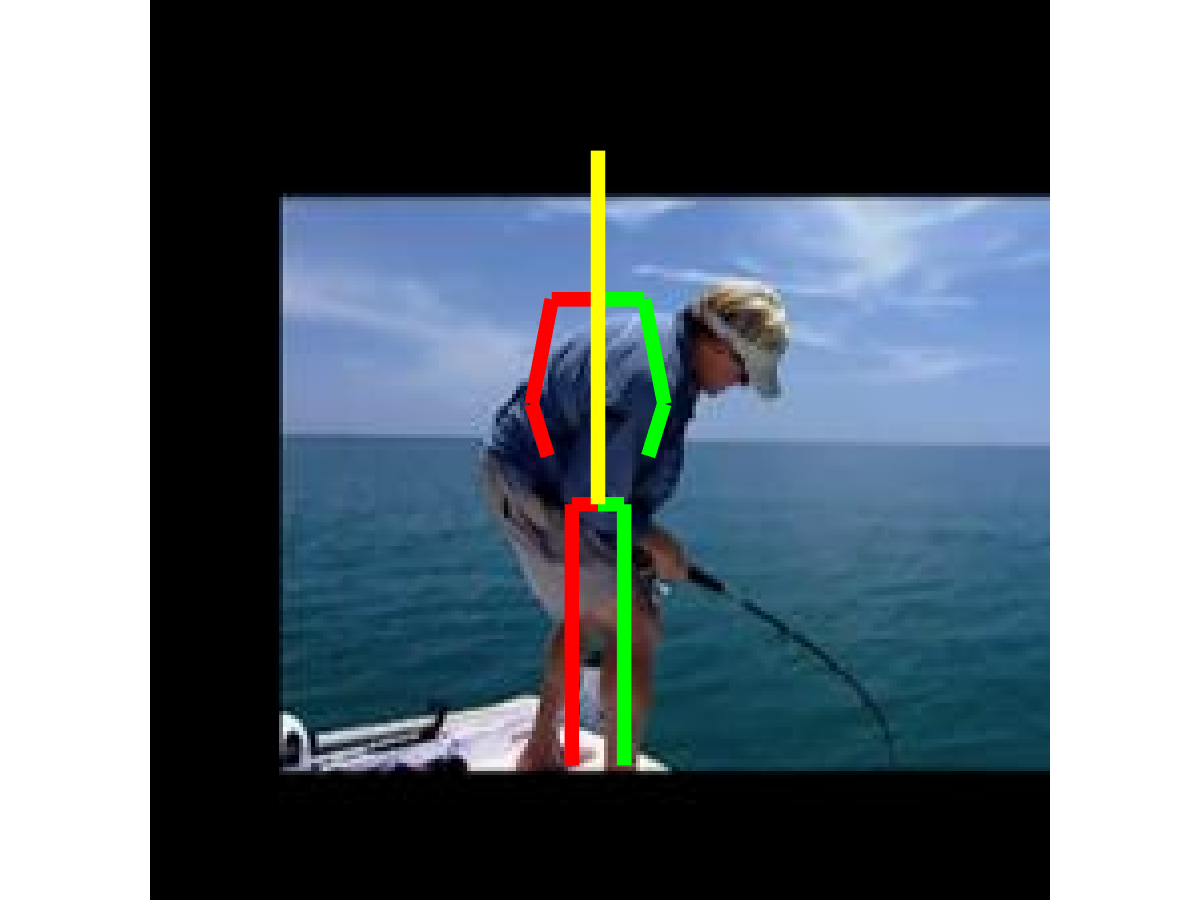} &
\includegraphics[height = 0.1\textheight, width = 0.2\textwidth, keepaspectratio = true]{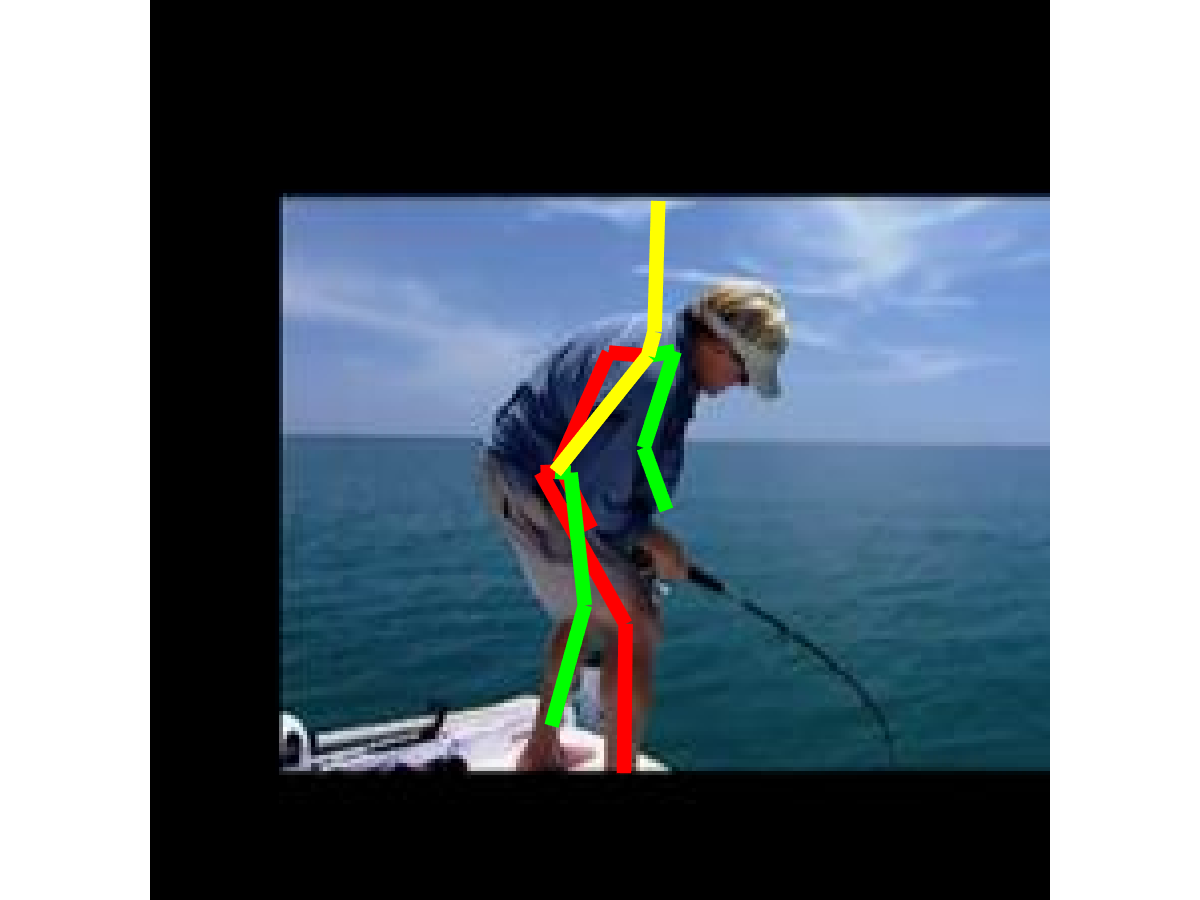} &
\includegraphics[height = 0.1\textheight, width = 0.2\textwidth, keepaspectratio = true]{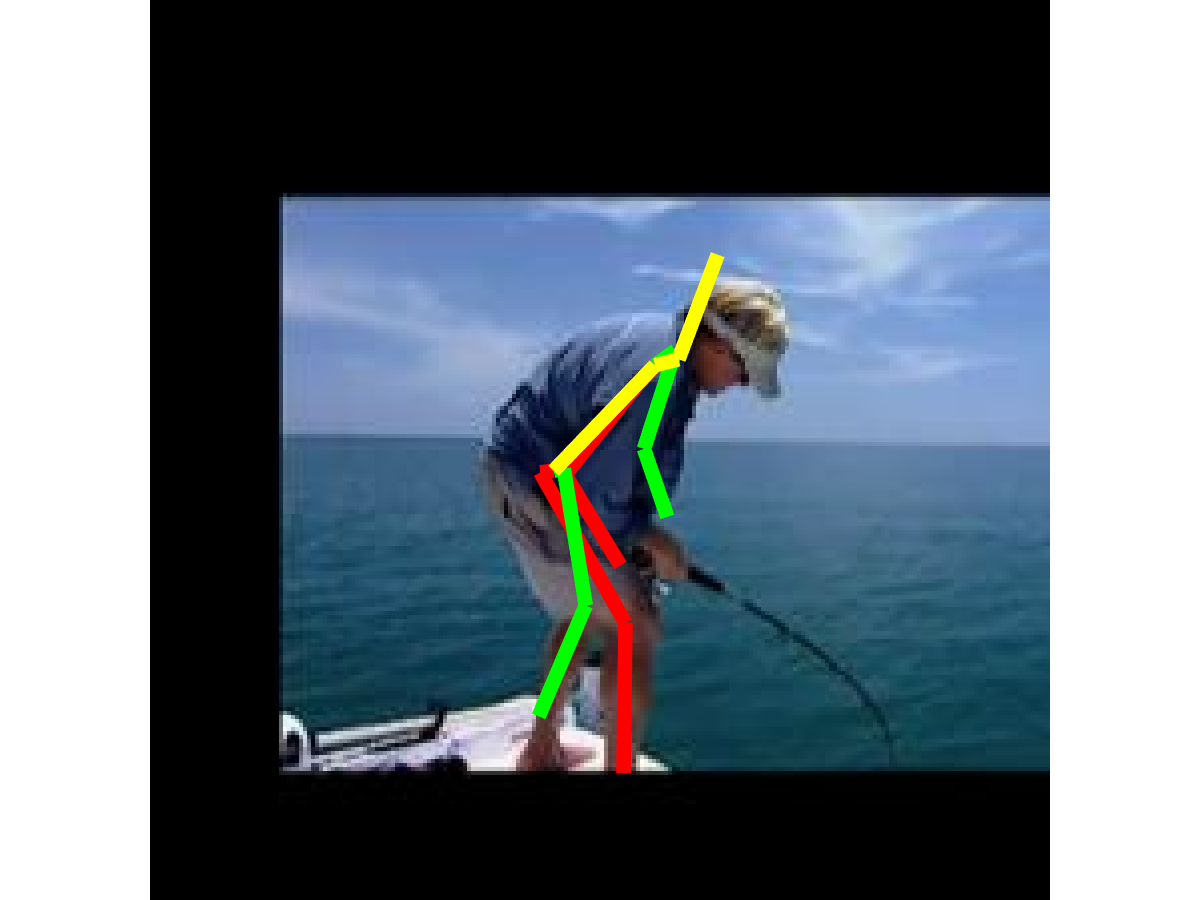} & 
\includegraphics[height = 0.1\textheight, width = 0.2\textwidth, keepaspectratio = true]{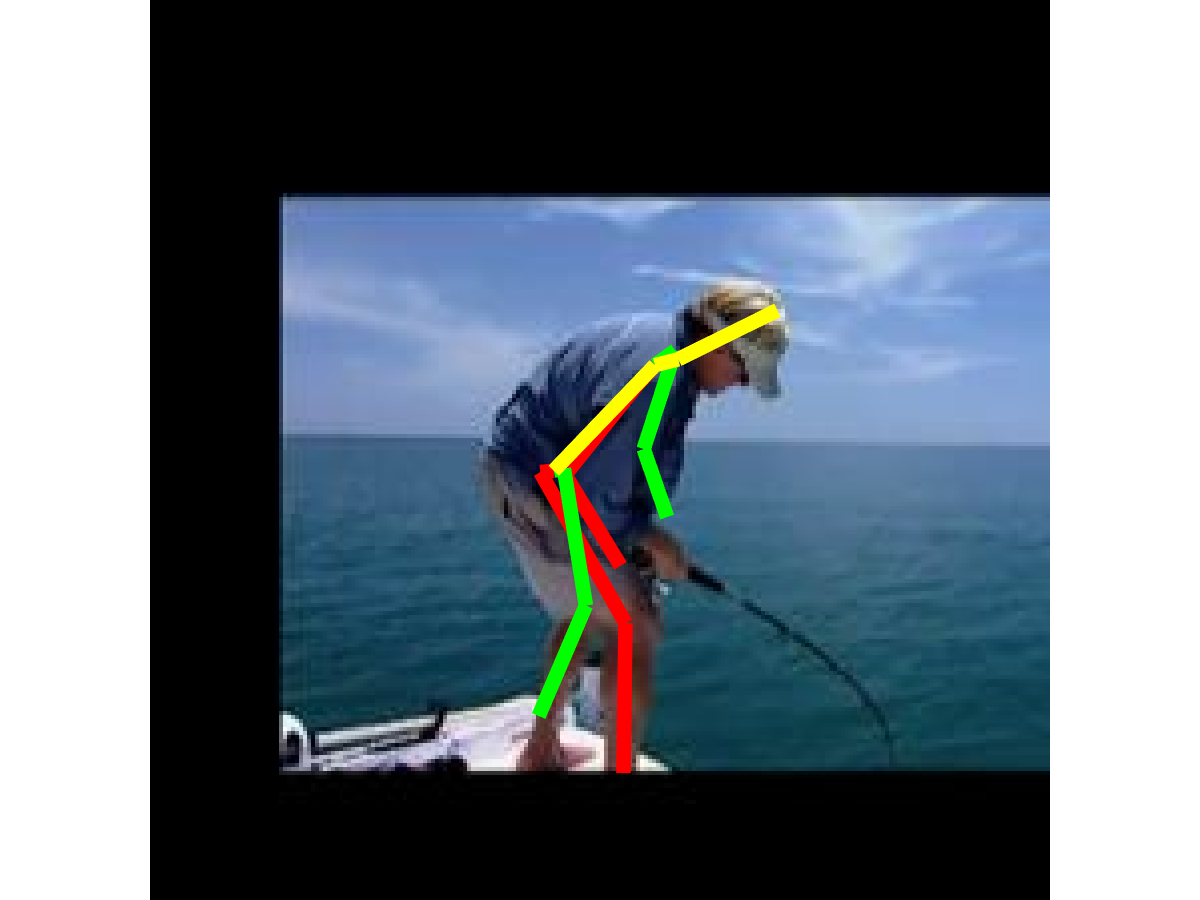} & 
\includegraphics[height = 0.1\textheight, width = 0.2\textwidth, keepaspectratio = true]{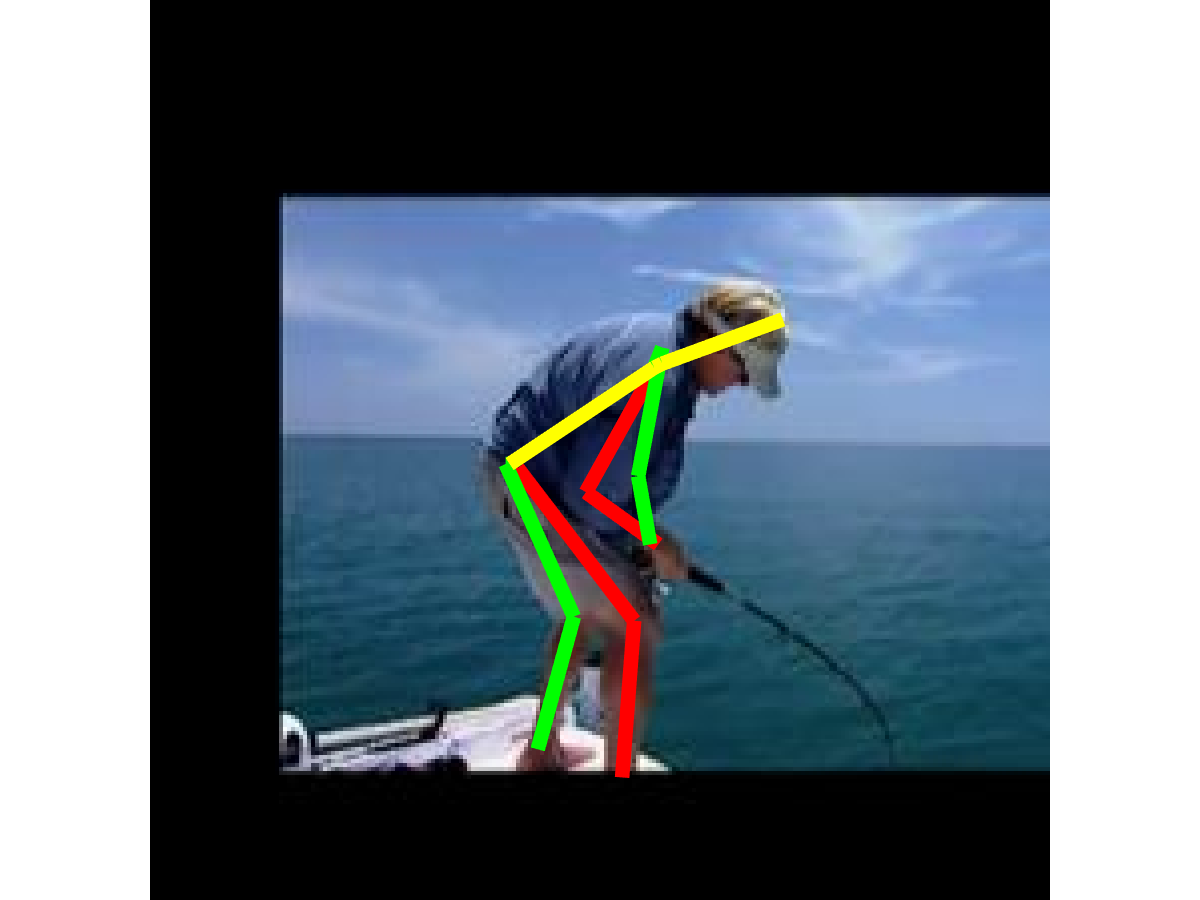} \\
\includegraphics[height = 0.1\textheight, width = 0.2\textwidth, keepaspectratio = true]{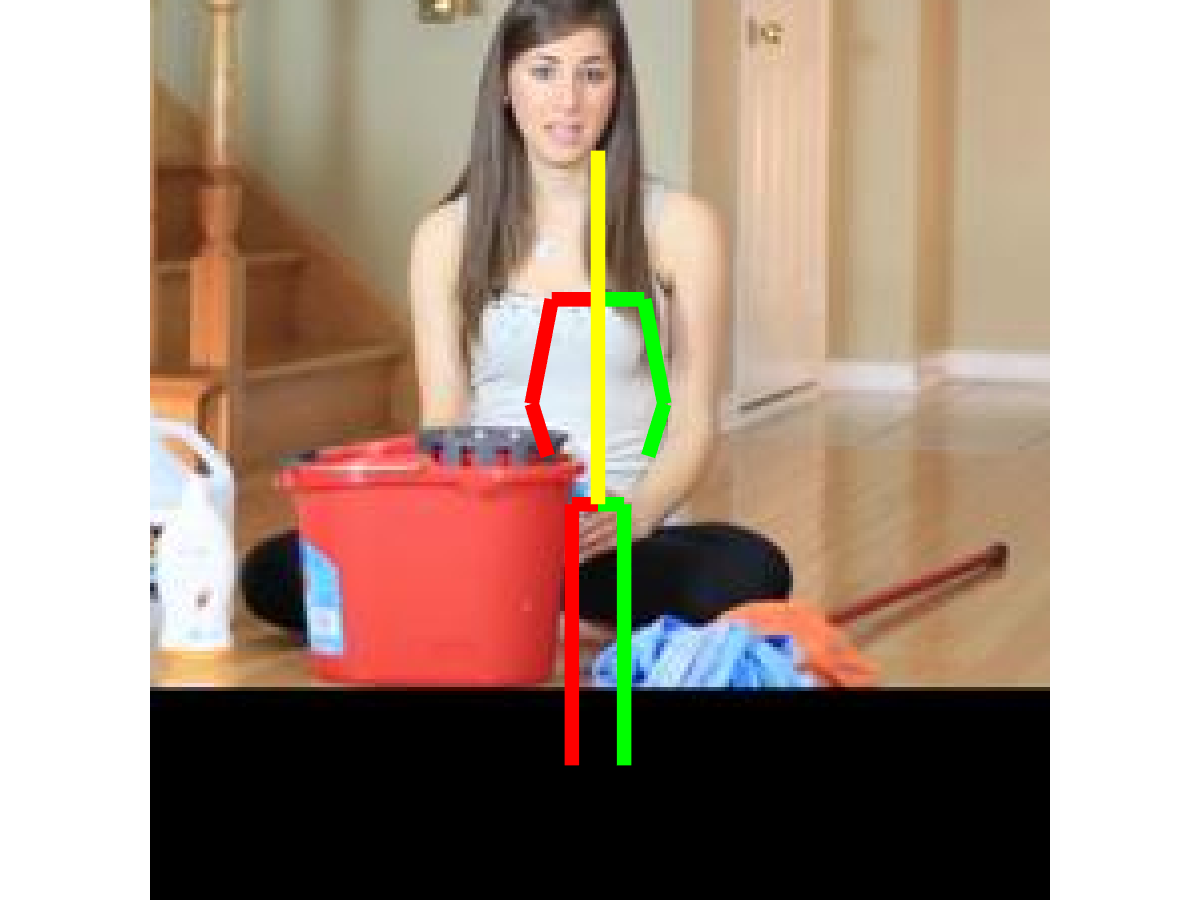} &
\includegraphics[height = 0.1\textheight, width = 0.2\textwidth, keepaspectratio = true]{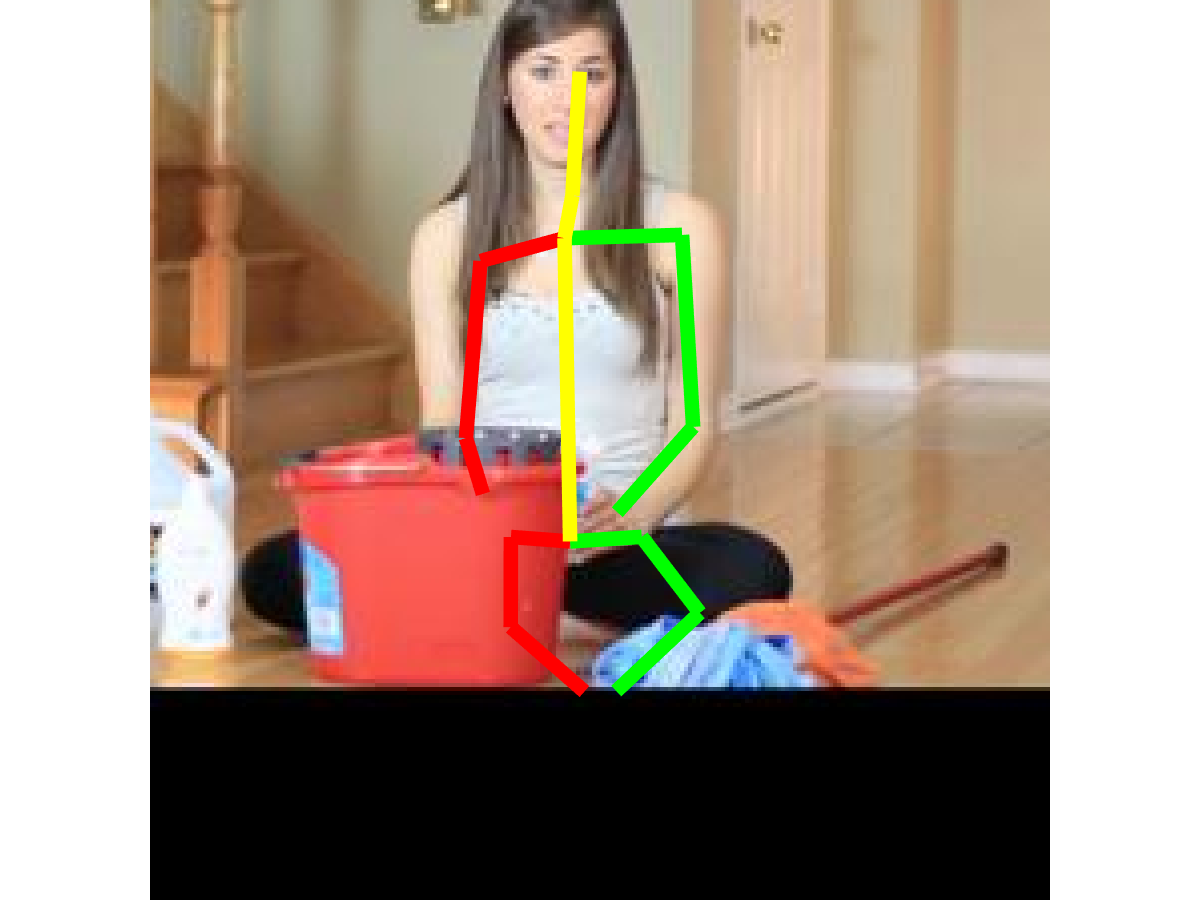} &
\includegraphics[height = 0.1\textheight, width = 0.2\textwidth, keepaspectratio = true]{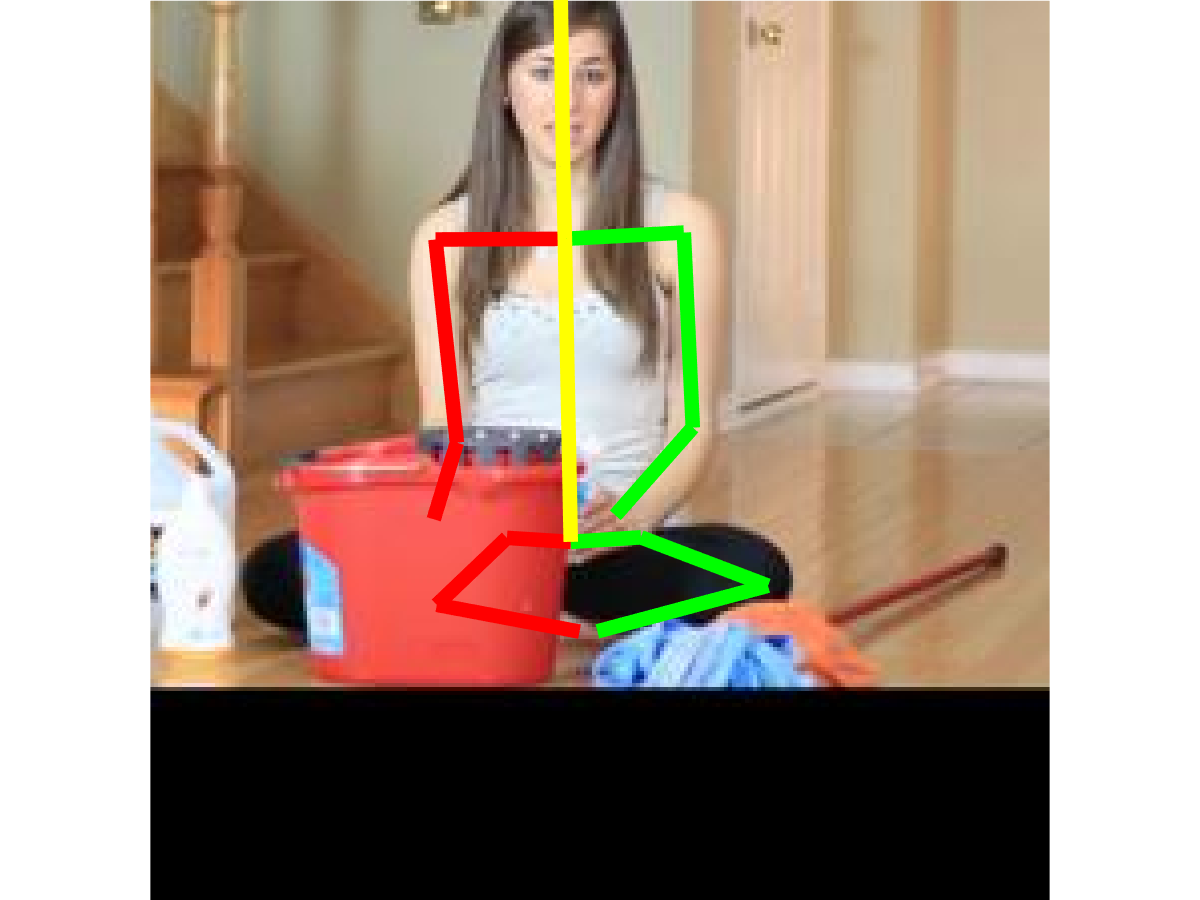} &
\includegraphics[height = 0.1\textheight, width = 0.2\textwidth, keepaspectratio = true]{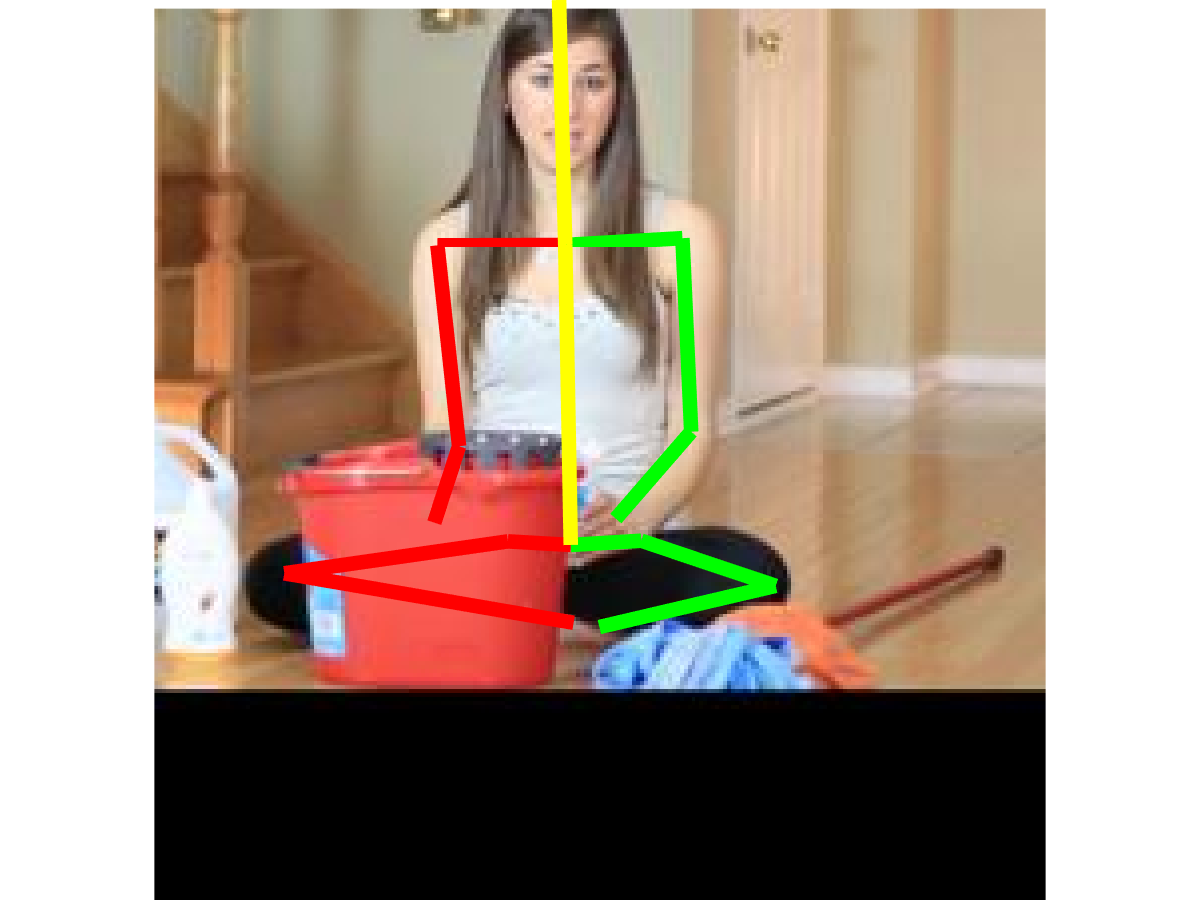} & 
\includegraphics[height = 0.1\textheight, width = 0.2\textwidth, keepaspectratio = true]{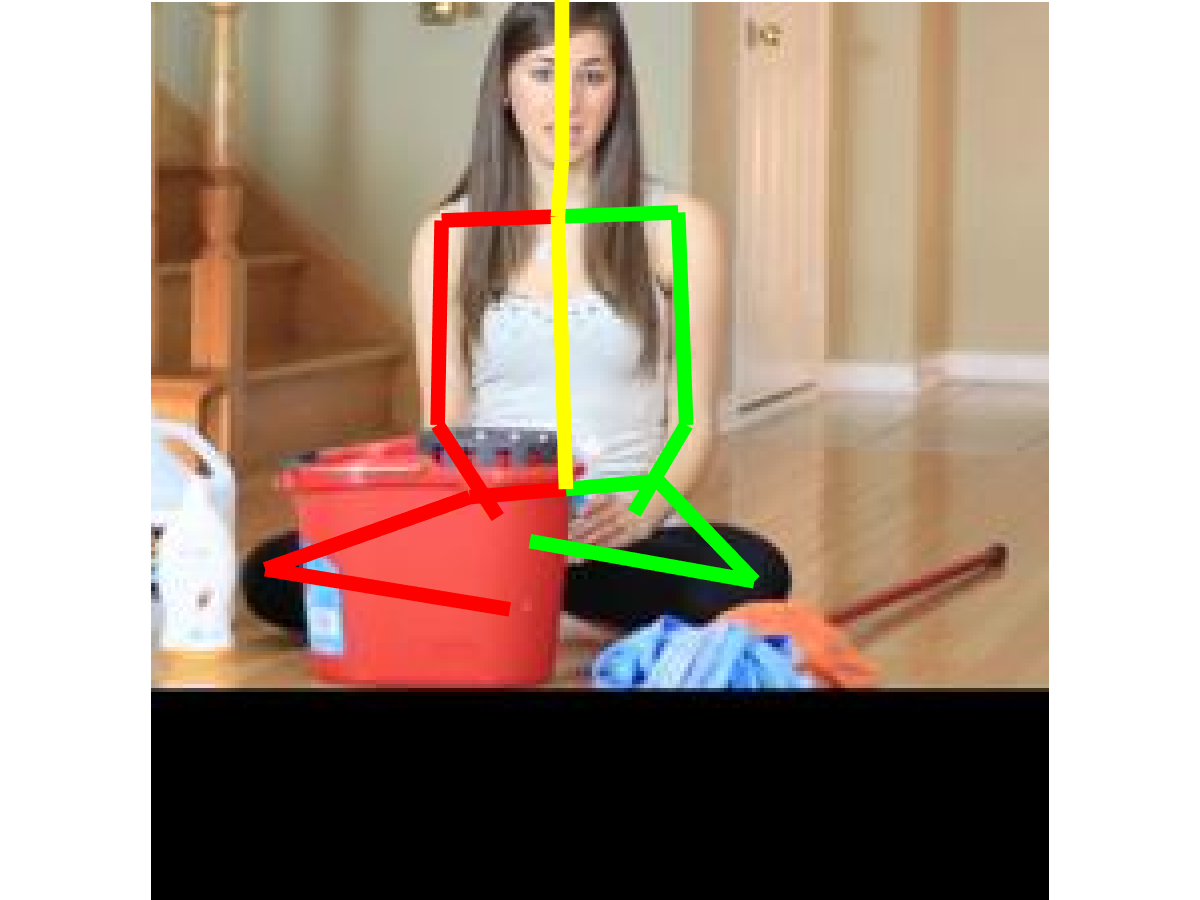} \\
\includegraphics[height = 0.1\textheight, width = 0.2\textwidth, keepaspectratio = true]{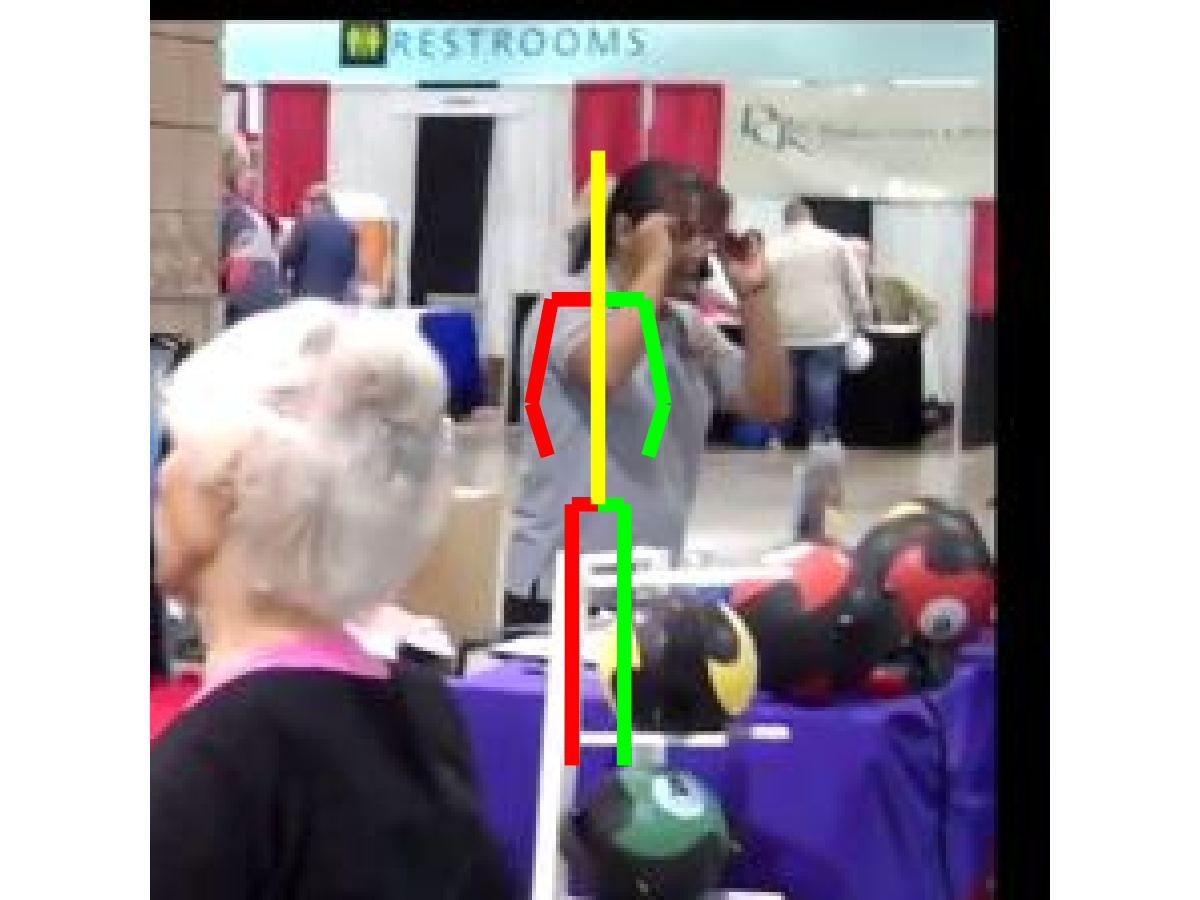} &
\includegraphics[height = 0.1\textheight, width = 0.2\textwidth, keepaspectratio = true]{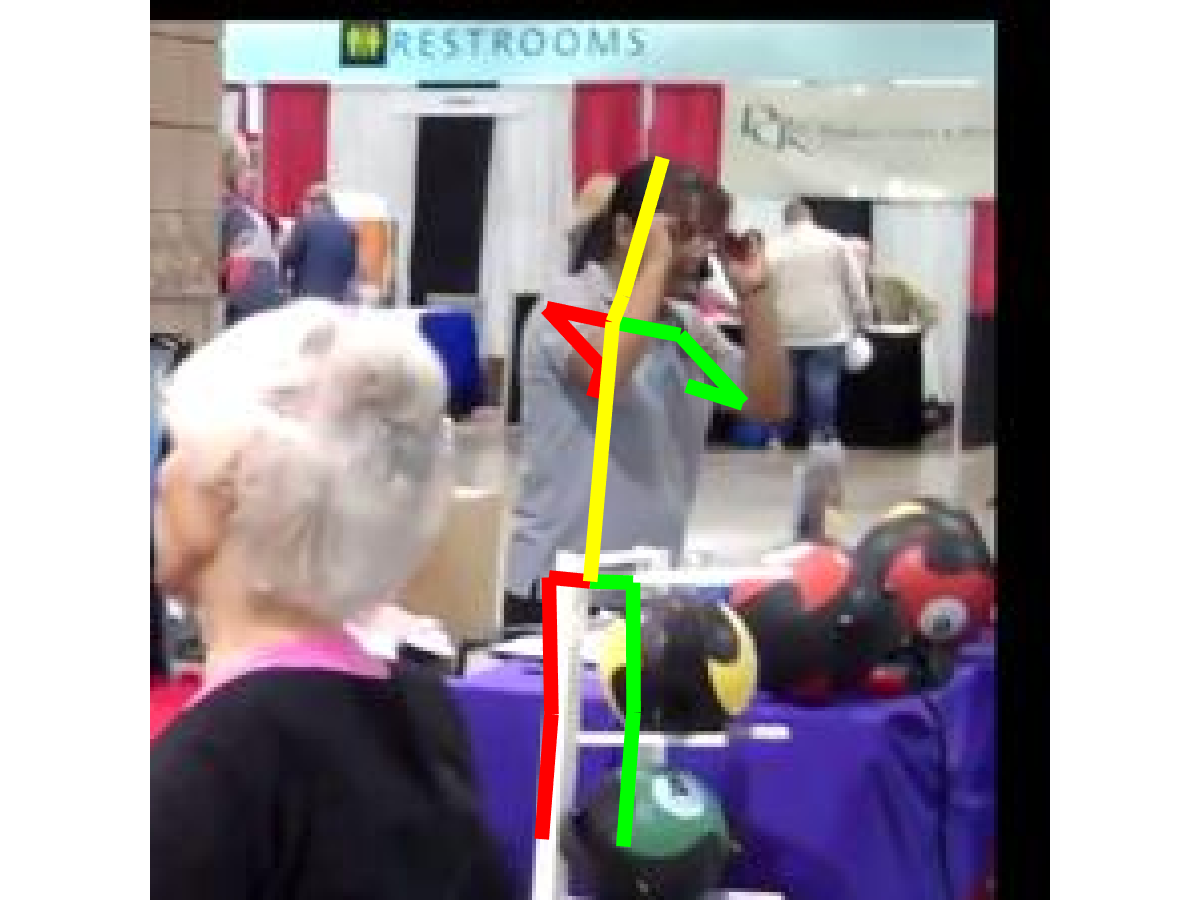} &
\includegraphics[height = 0.1\textheight, width = 0.2\textwidth, keepaspectratio = true]{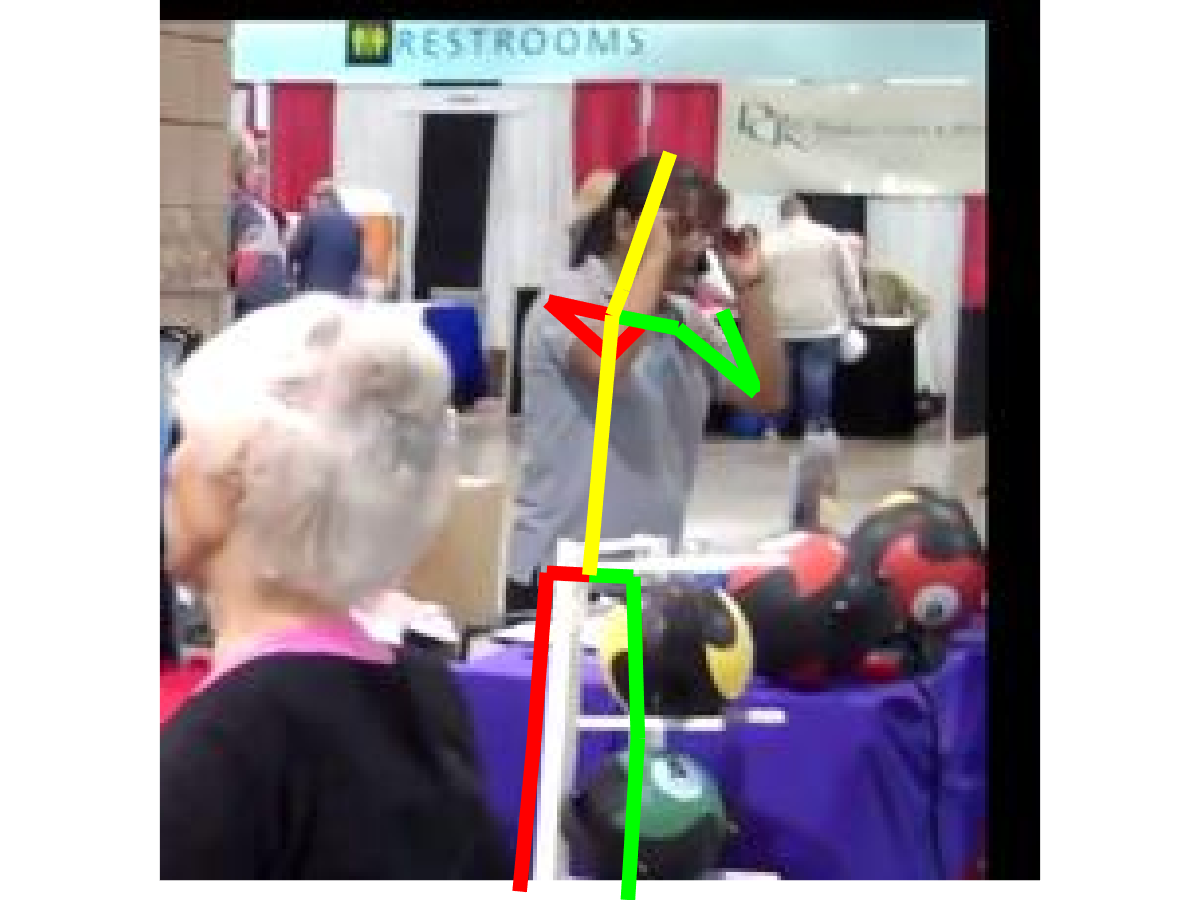} & 
\includegraphics[height = 0.1\textheight, width = 0.2\textwidth, keepaspectratio = true]{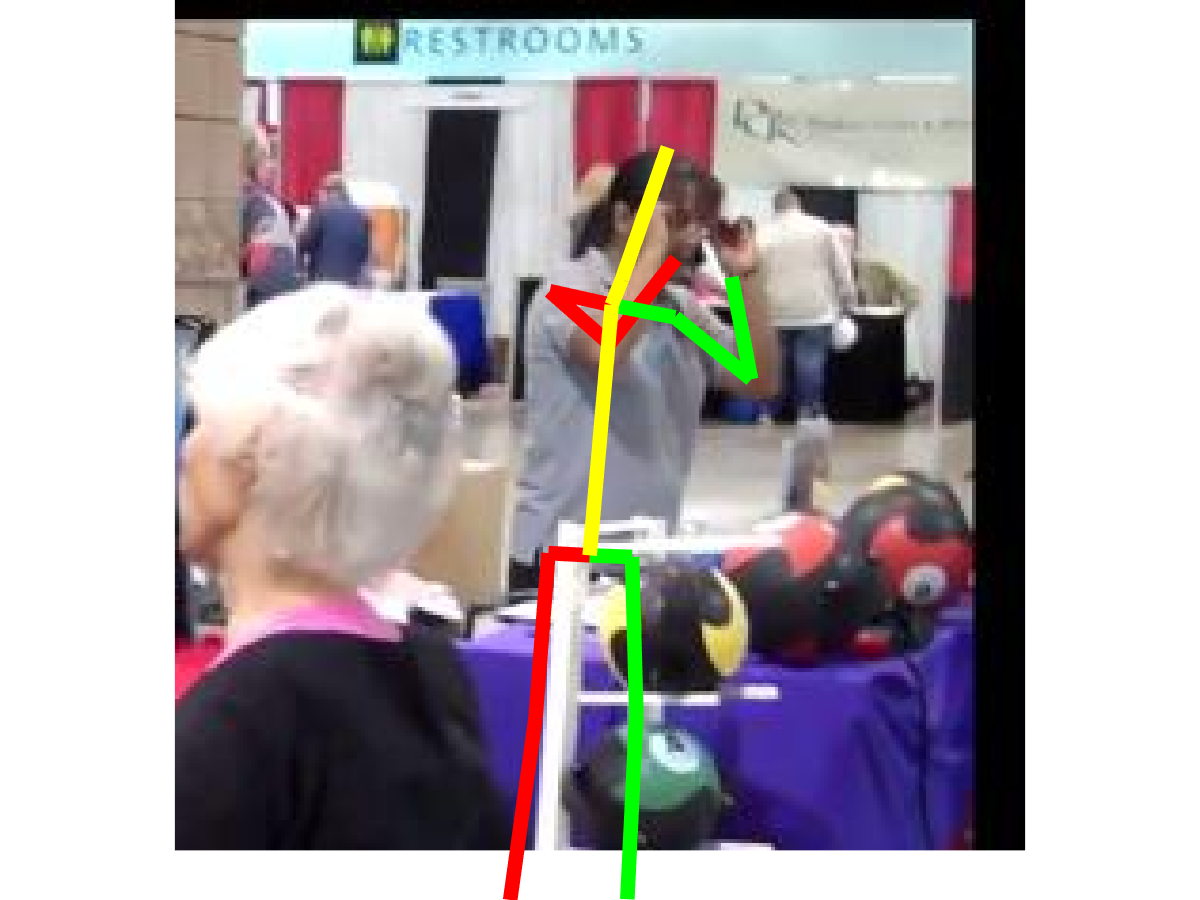} & 
\includegraphics[height = 0.1\textheight, width = 0.2\textwidth, keepaspectratio = true]{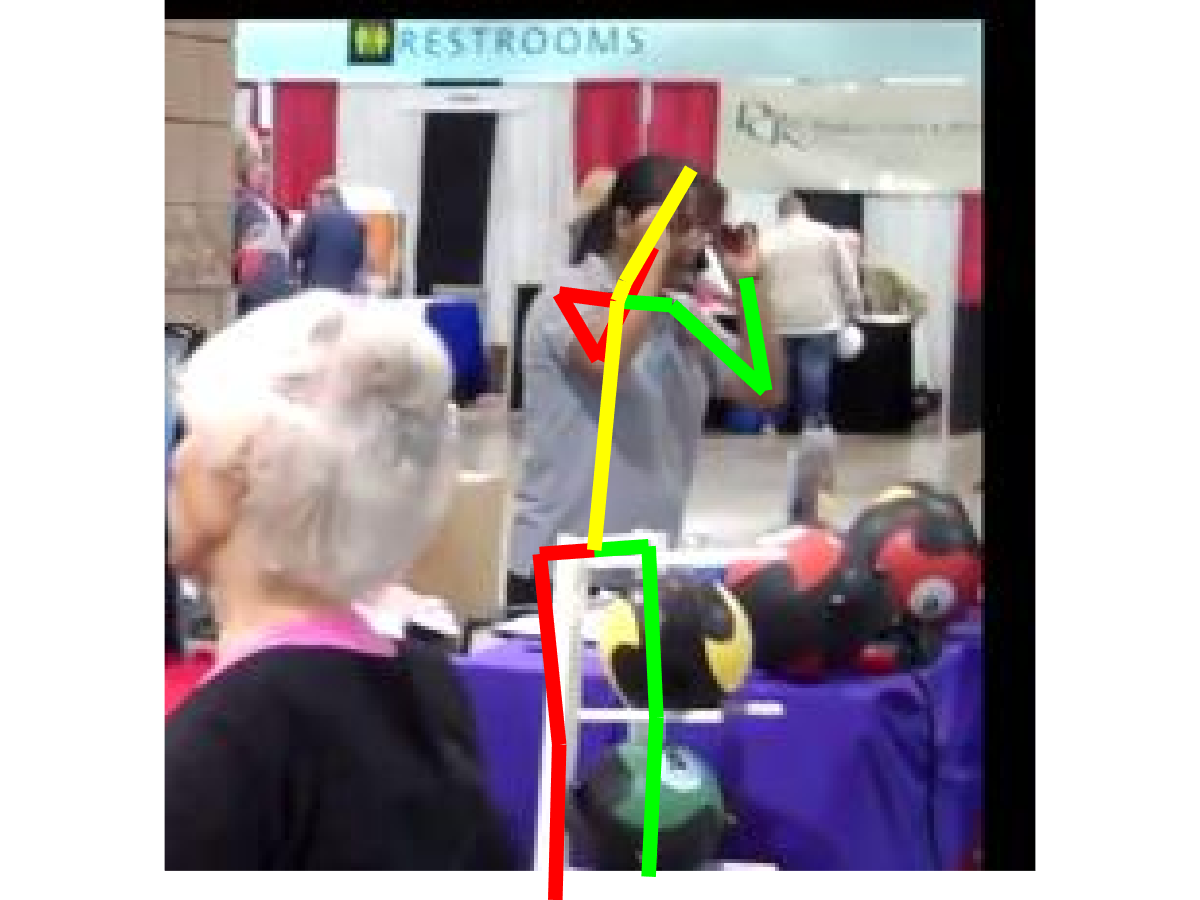} \\
\includegraphics[height = 0.1\textheight, width = 0.2\textwidth, keepaspectratio = true]{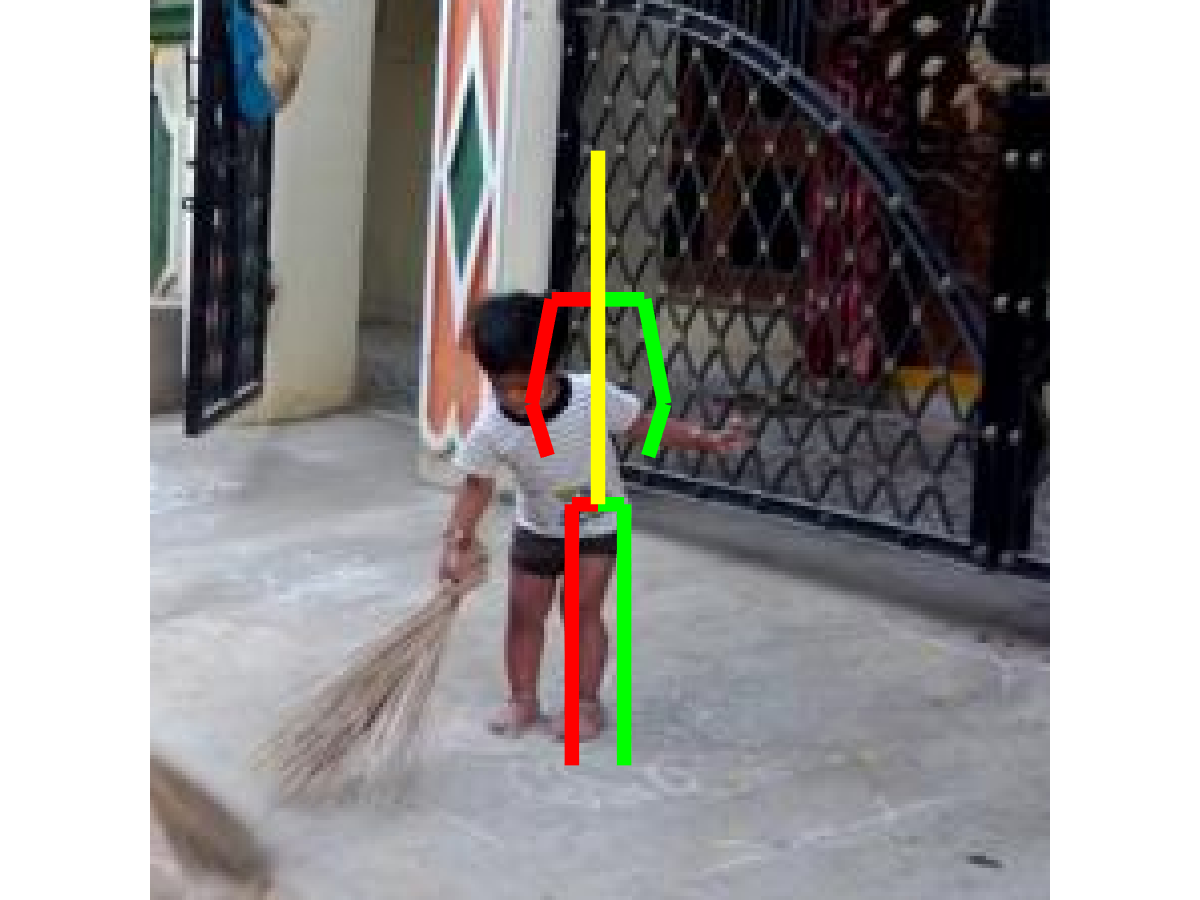} &
\includegraphics[height = 0.1\textheight, width = 0.2\textwidth, keepaspectratio = true]{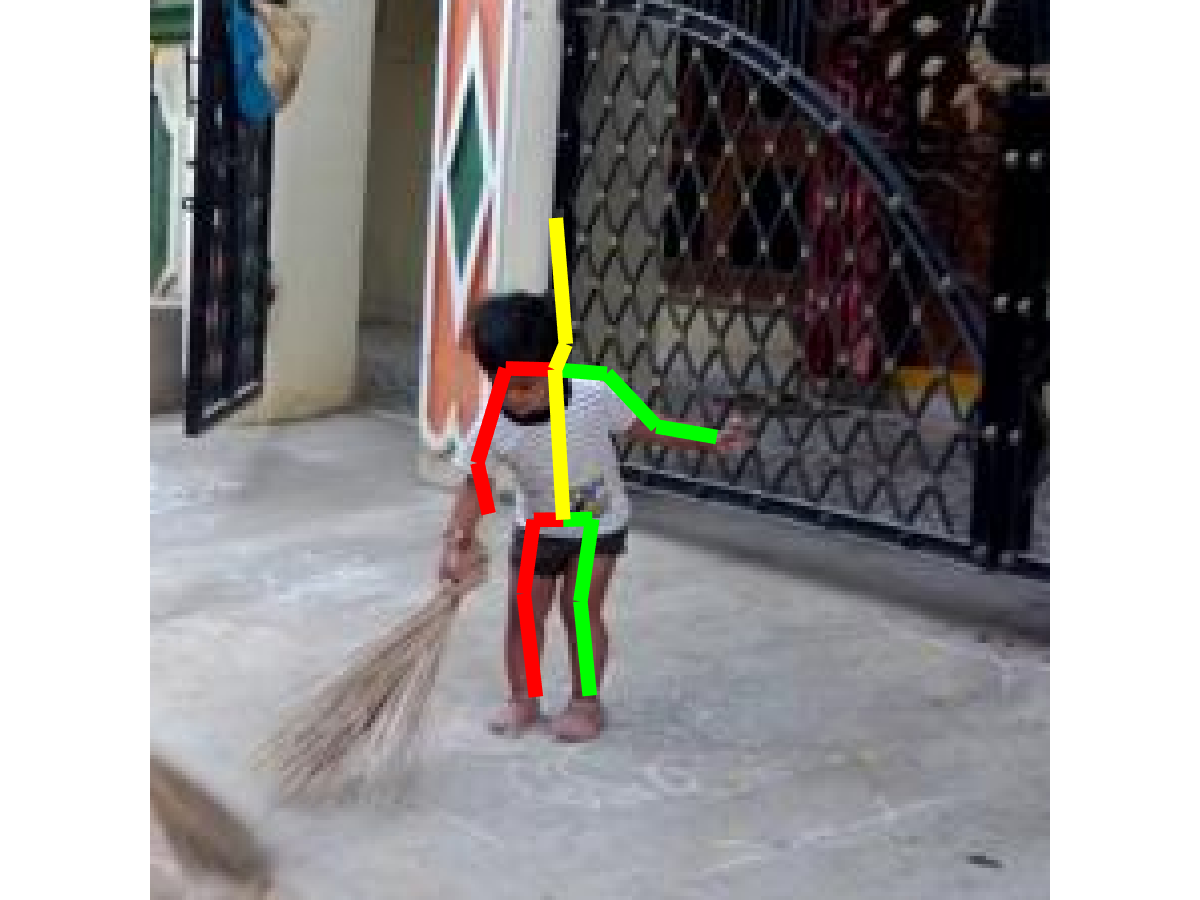} &
\includegraphics[height = 0.1\textheight, width = 0.2\textwidth, keepaspectratio = true]{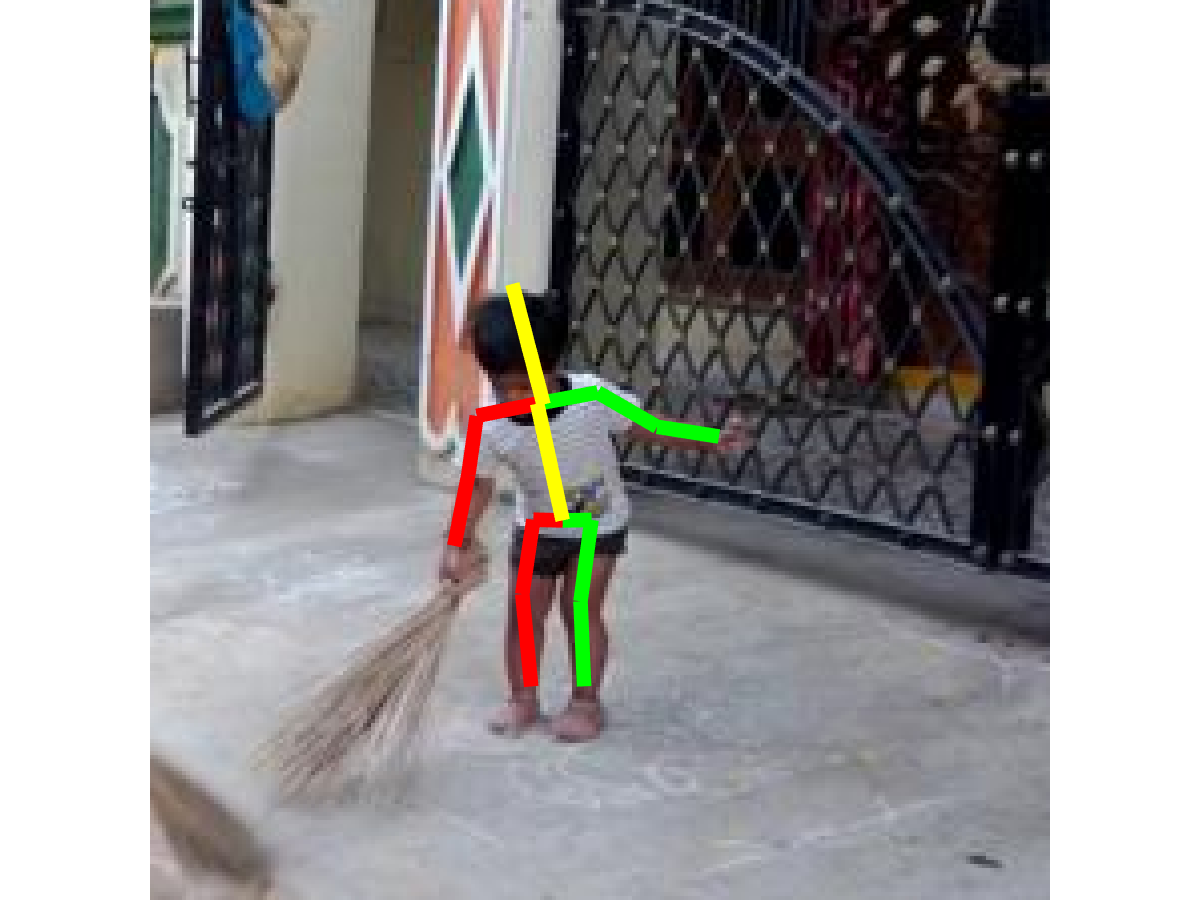} & 
\includegraphics[height = 0.1\textheight, width = 0.2\textwidth, keepaspectratio = true]{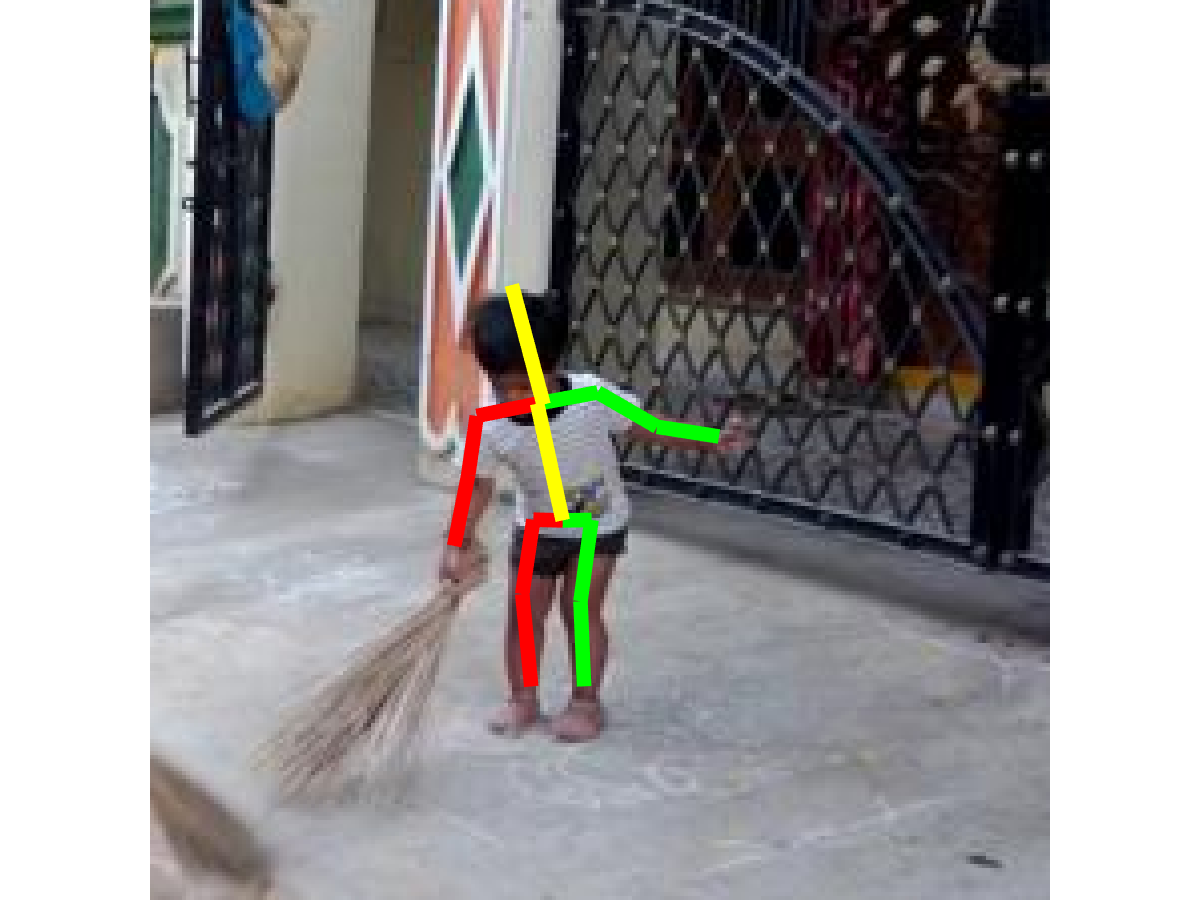} & 
\includegraphics[height = 0.1\textheight, width = 0.2\textwidth, keepaspectratio = true]{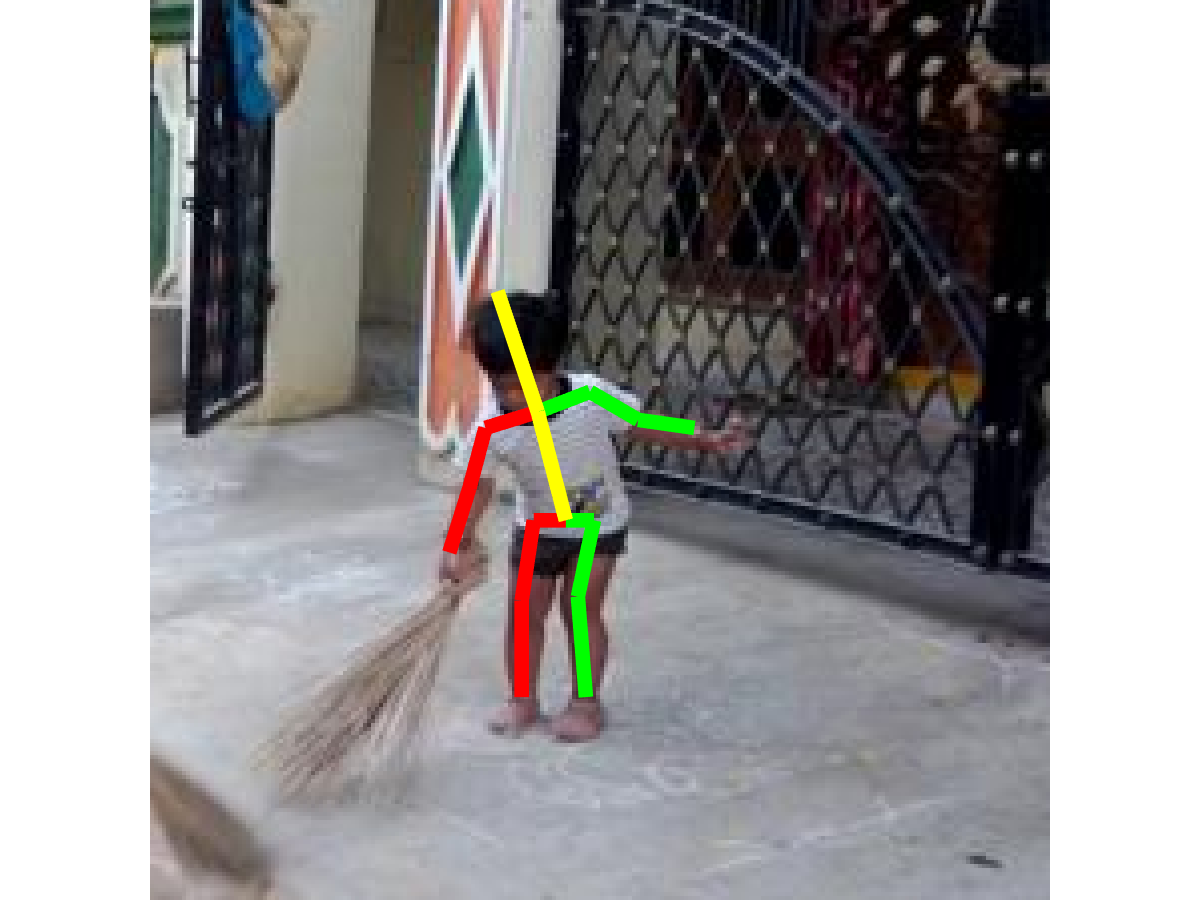} \\
\includegraphics[height = 0.1\textheight, width = 0.2\textwidth, keepaspectratio = true]{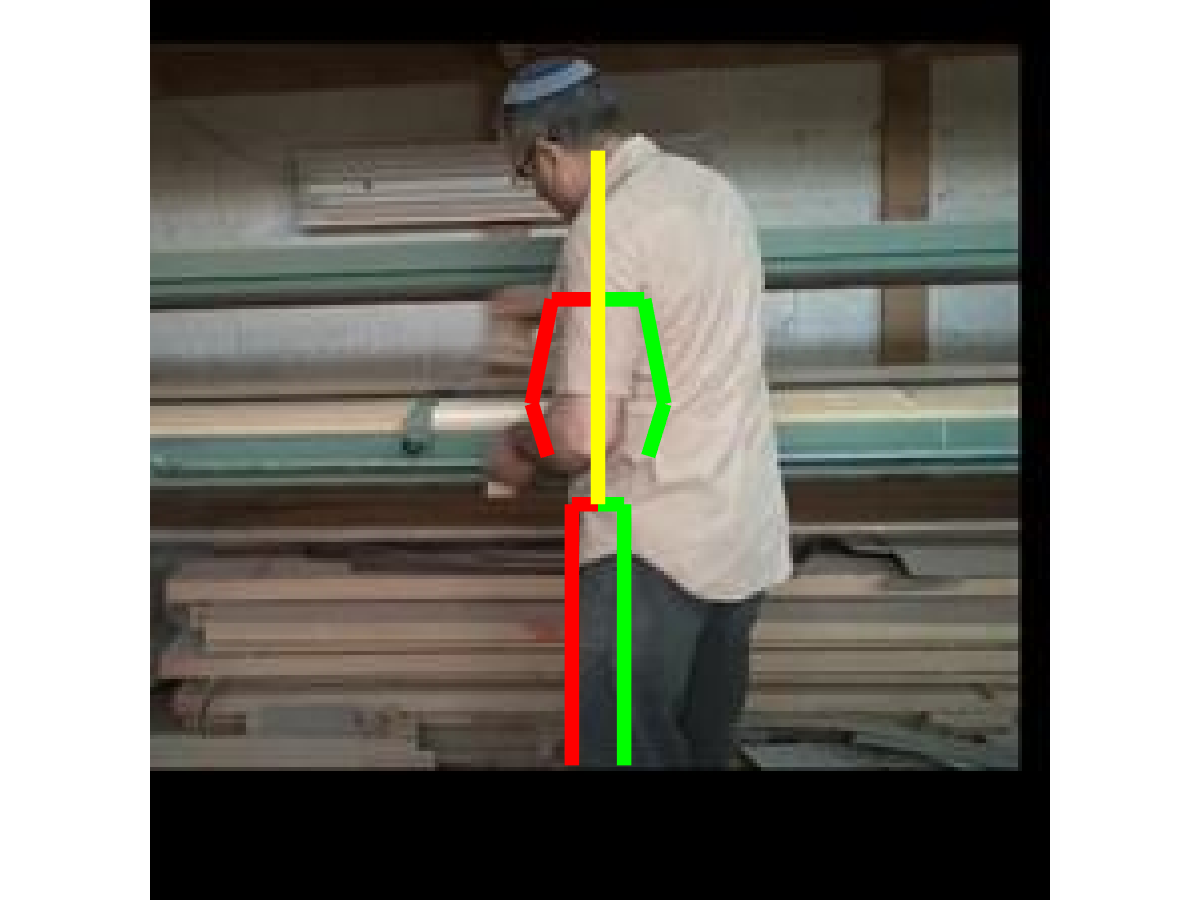} &
\includegraphics[height = 0.1\textheight, width = 0.2\textwidth, keepaspectratio = true]{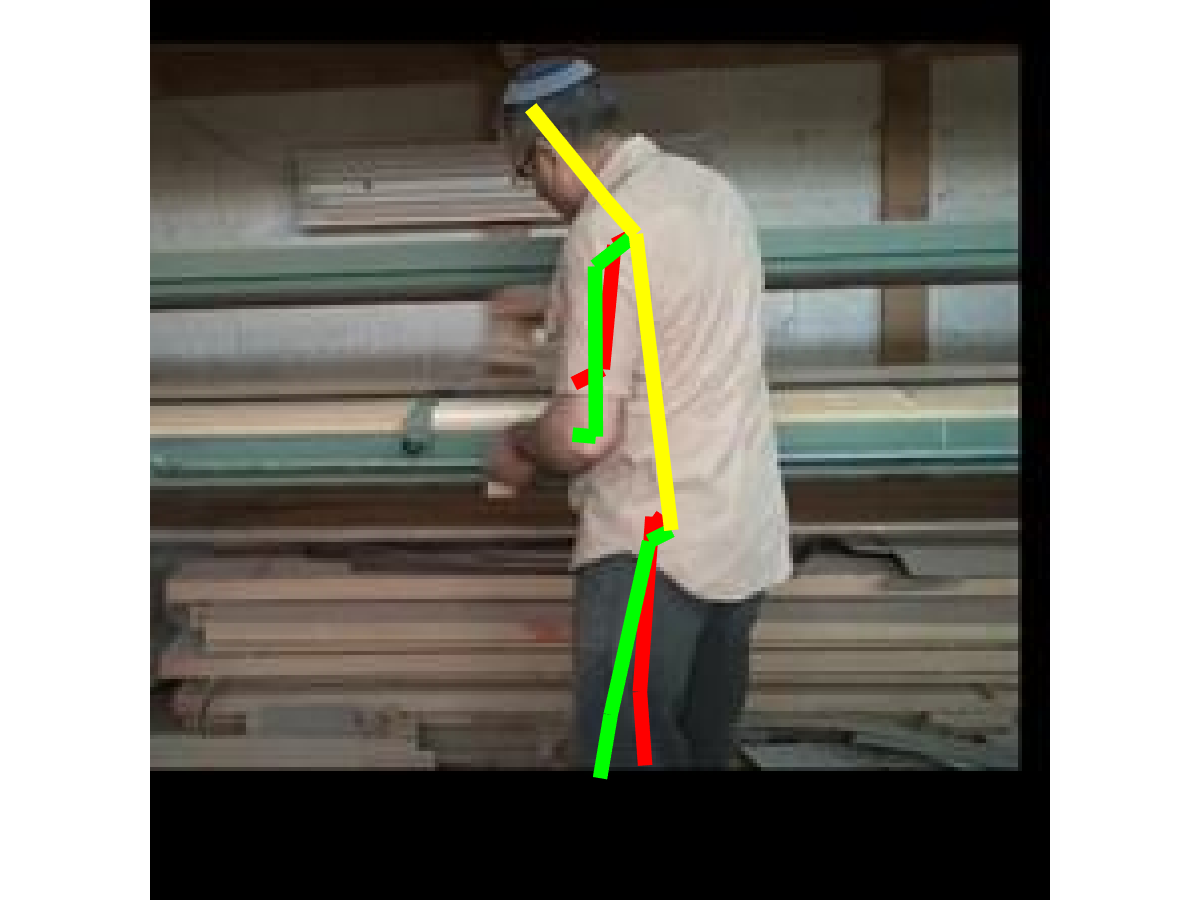} &
\includegraphics[height = 0.1\textheight, width = 0.2\textwidth, keepaspectratio = true]{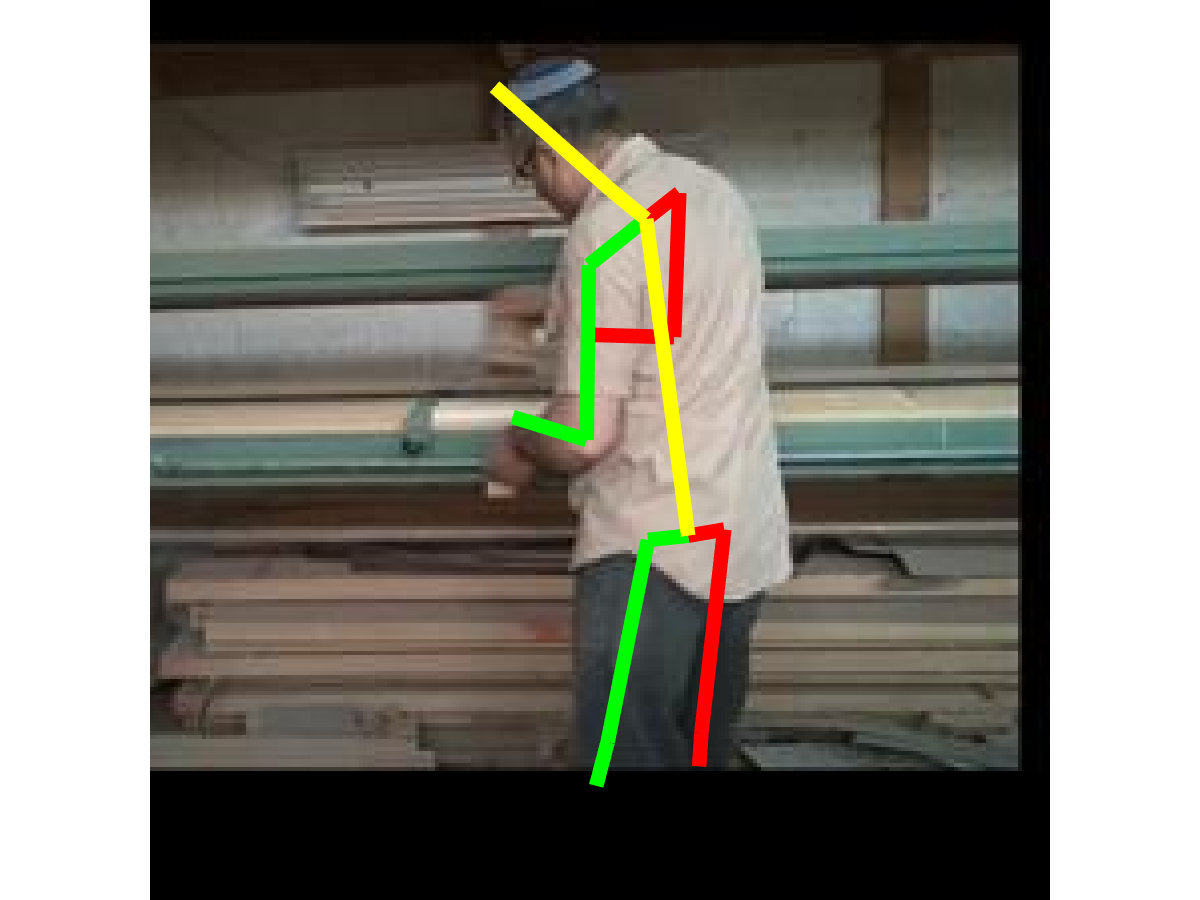} & 
\includegraphics[height = 0.1\textheight, width = 0.2\textwidth, keepaspectratio = true]{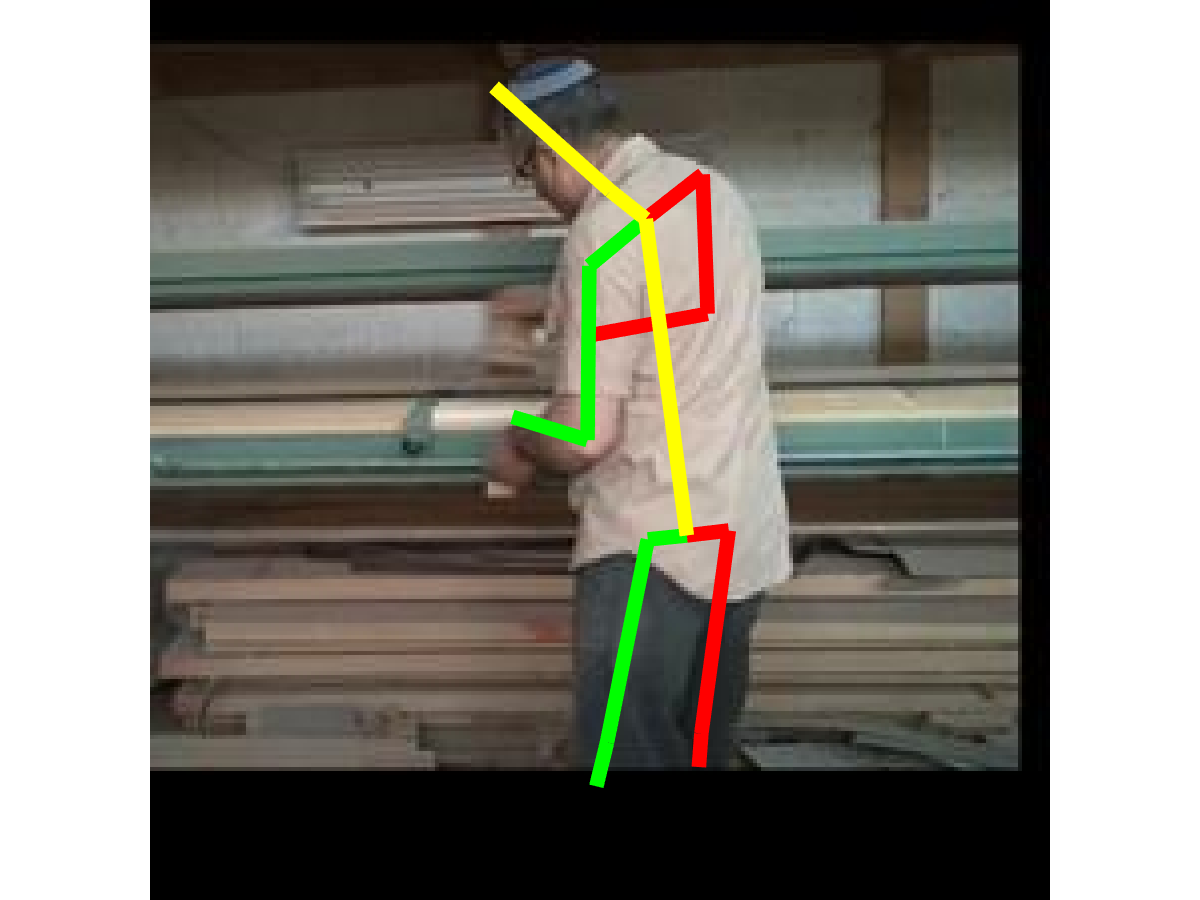} & 
\includegraphics[height = 0.1\textheight, width = 0.2\textwidth, keepaspectratio = true]{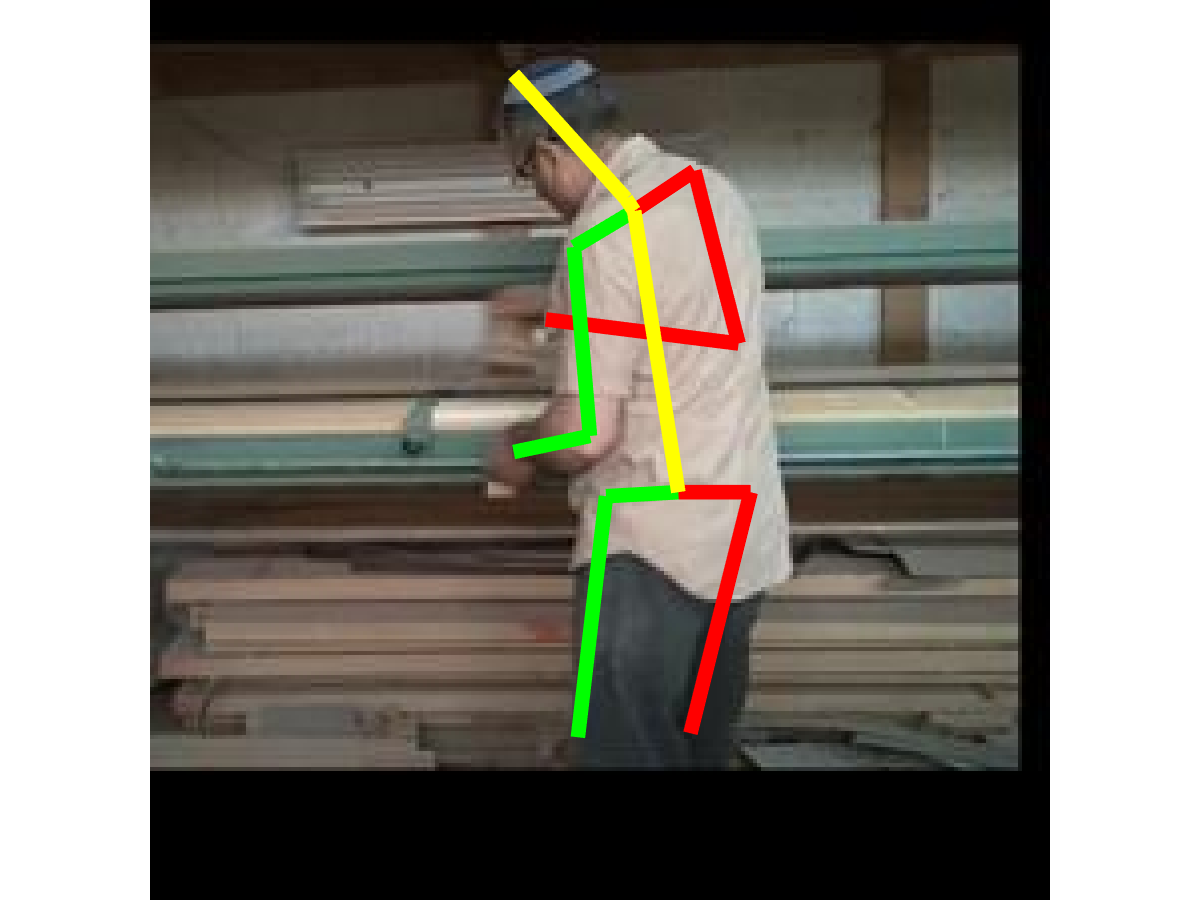} \\
\includegraphics[height = 0.1\textheight, width = 0.2\textwidth, keepaspectratio = true]{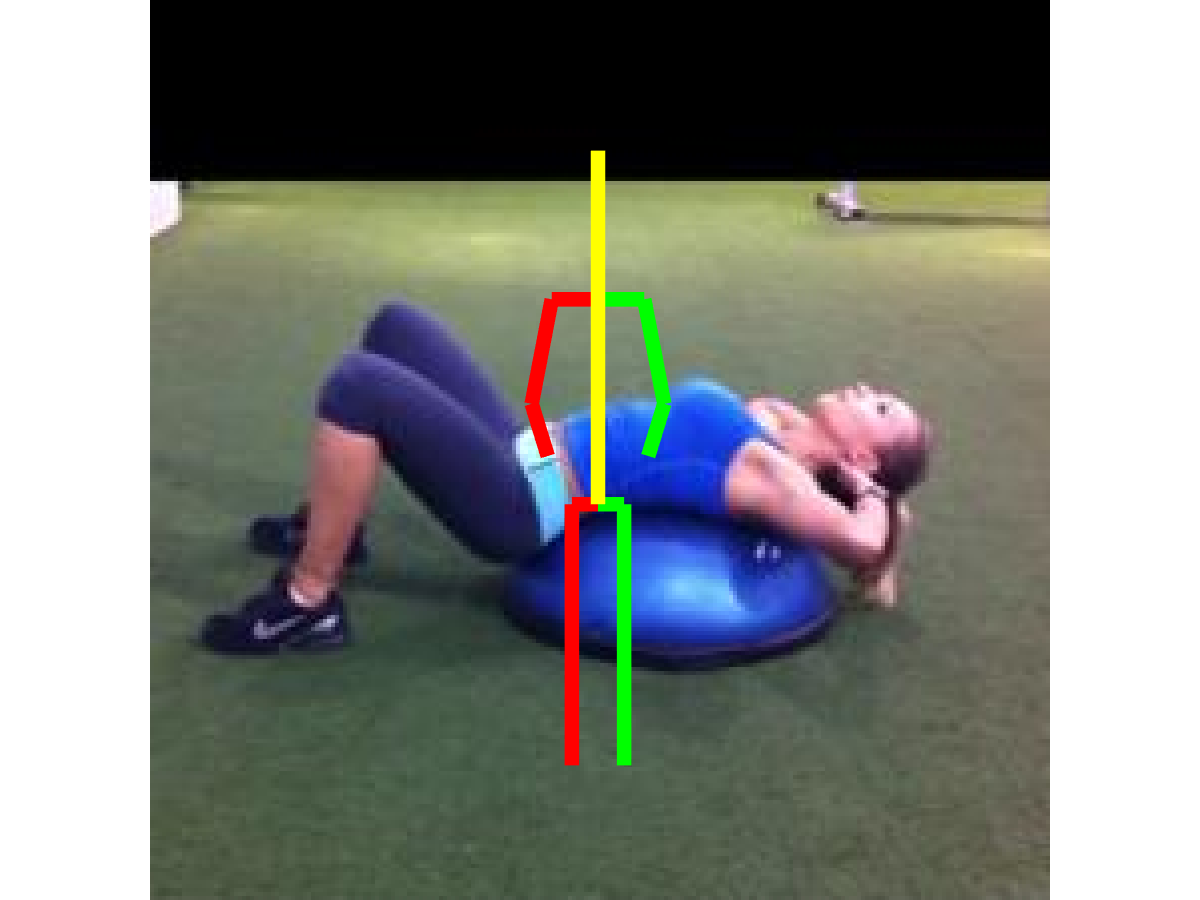} &
\includegraphics[height = 0.1\textheight, width = 0.2\textwidth, keepaspectratio = true]{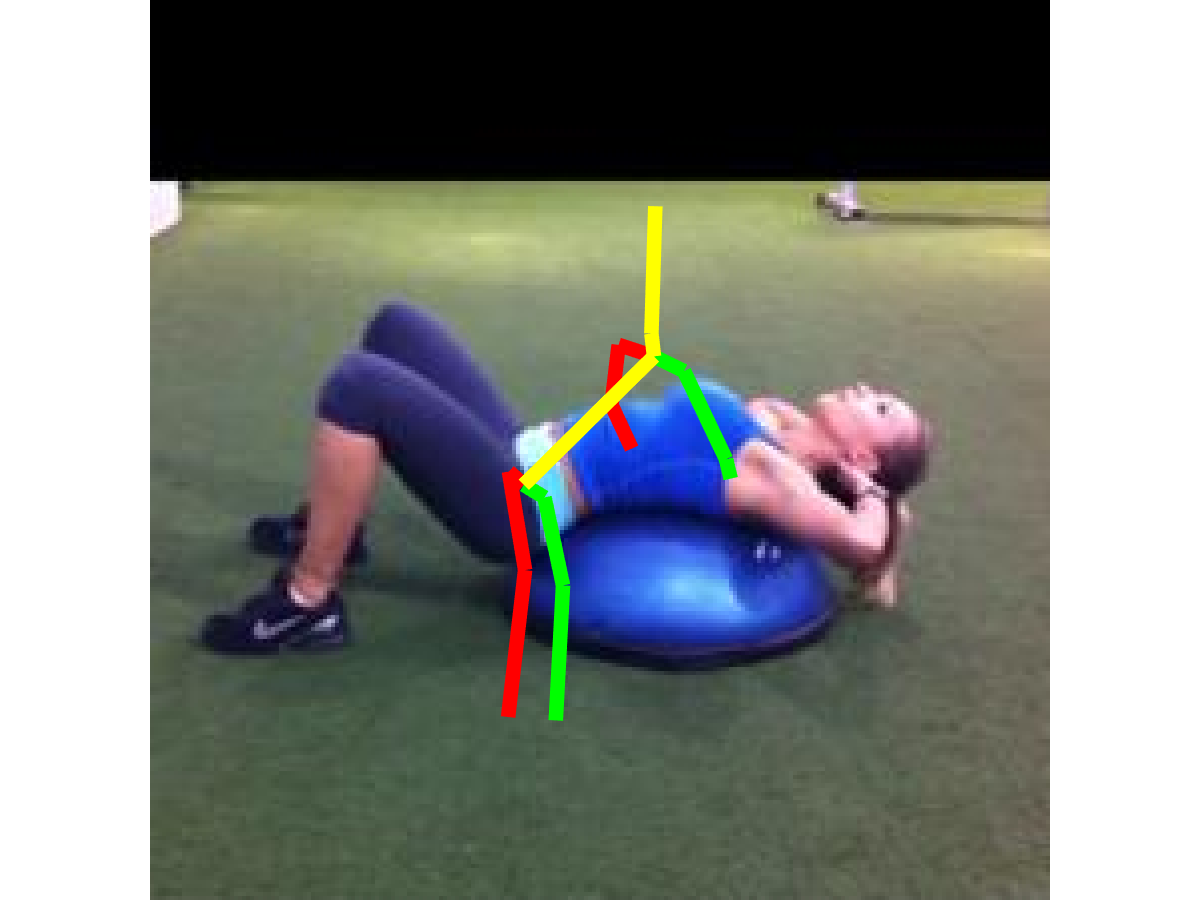} &
\includegraphics[height = 0.1\textheight, width = 0.2\textwidth, keepaspectratio = true]{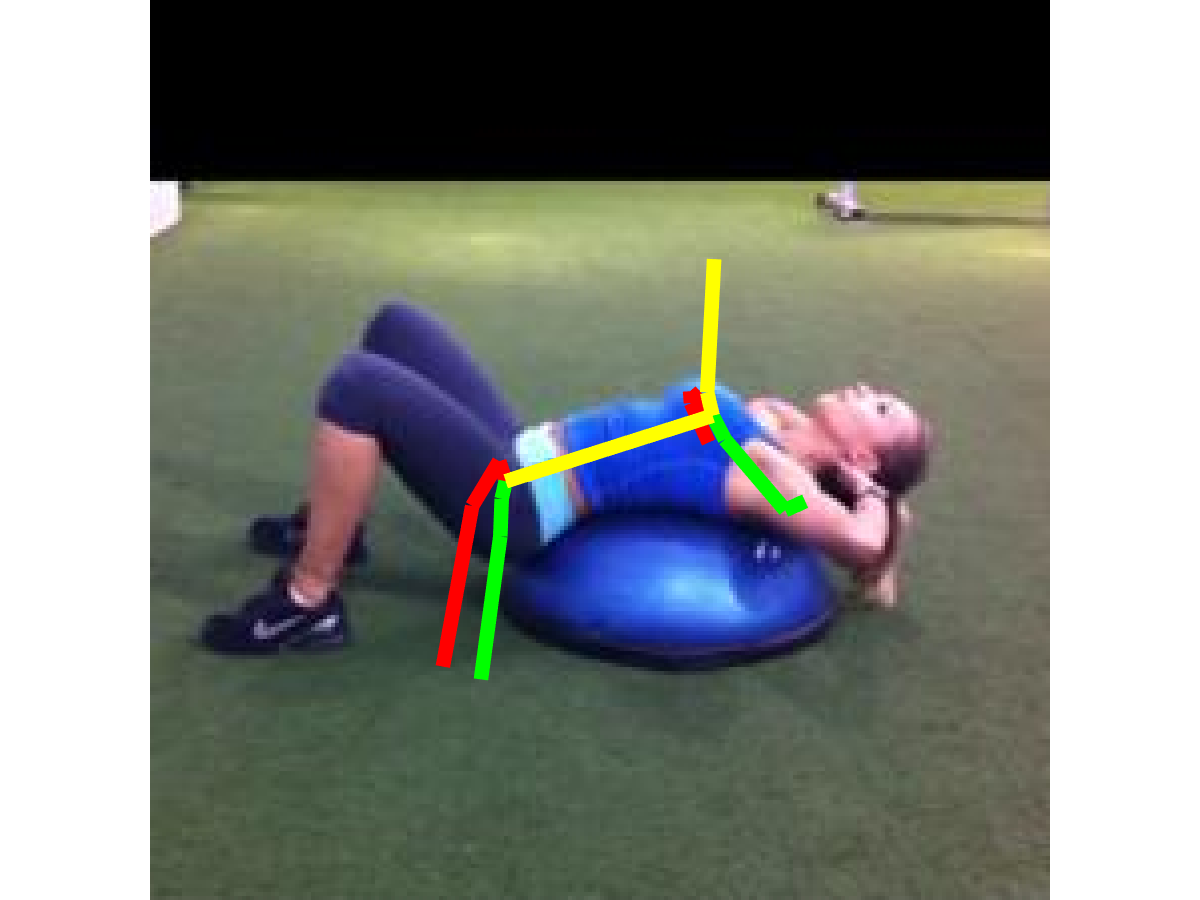} &
\includegraphics[height = 0.1\textheight, width = 0.2\textwidth, keepaspectratio = true]{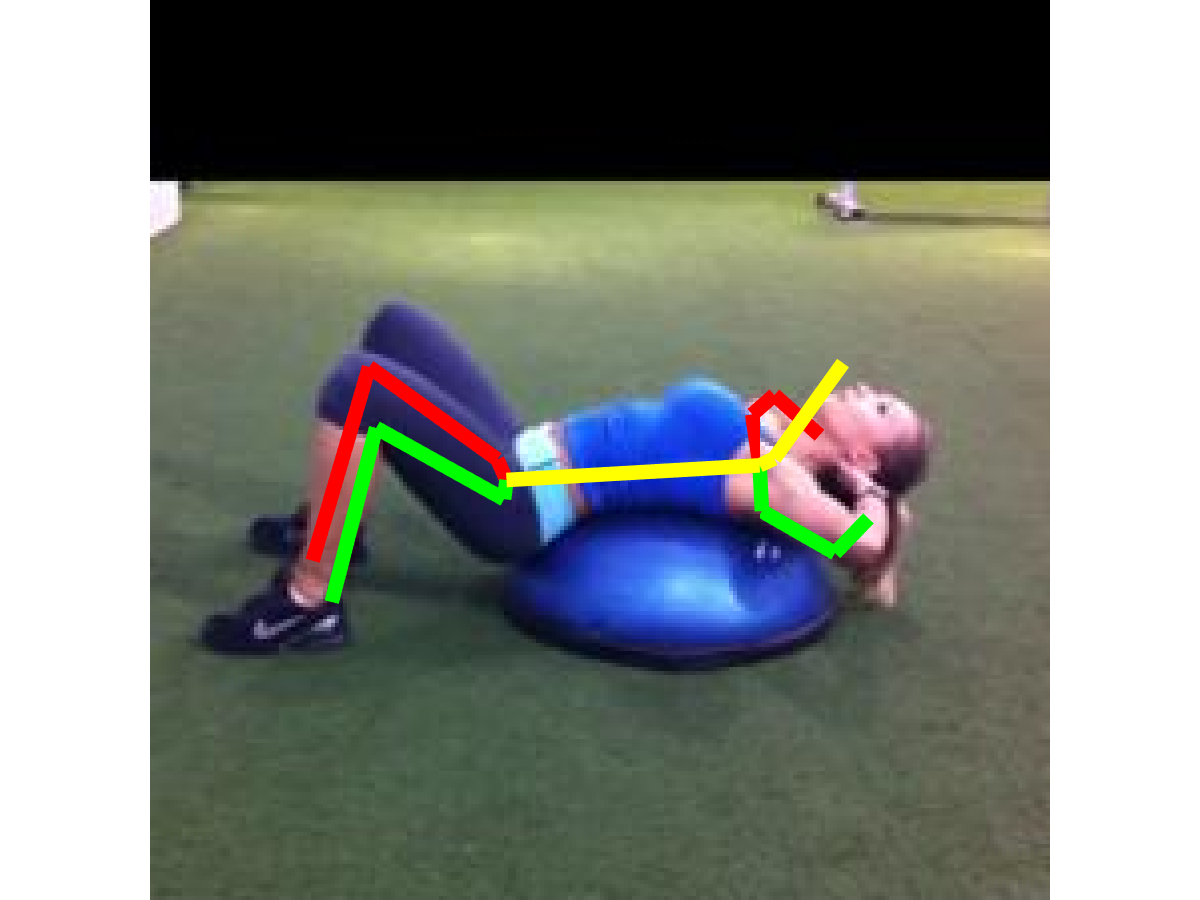} & 
\includegraphics[height = 0.1\textheight, width = 0.2\textwidth, keepaspectratio = true]{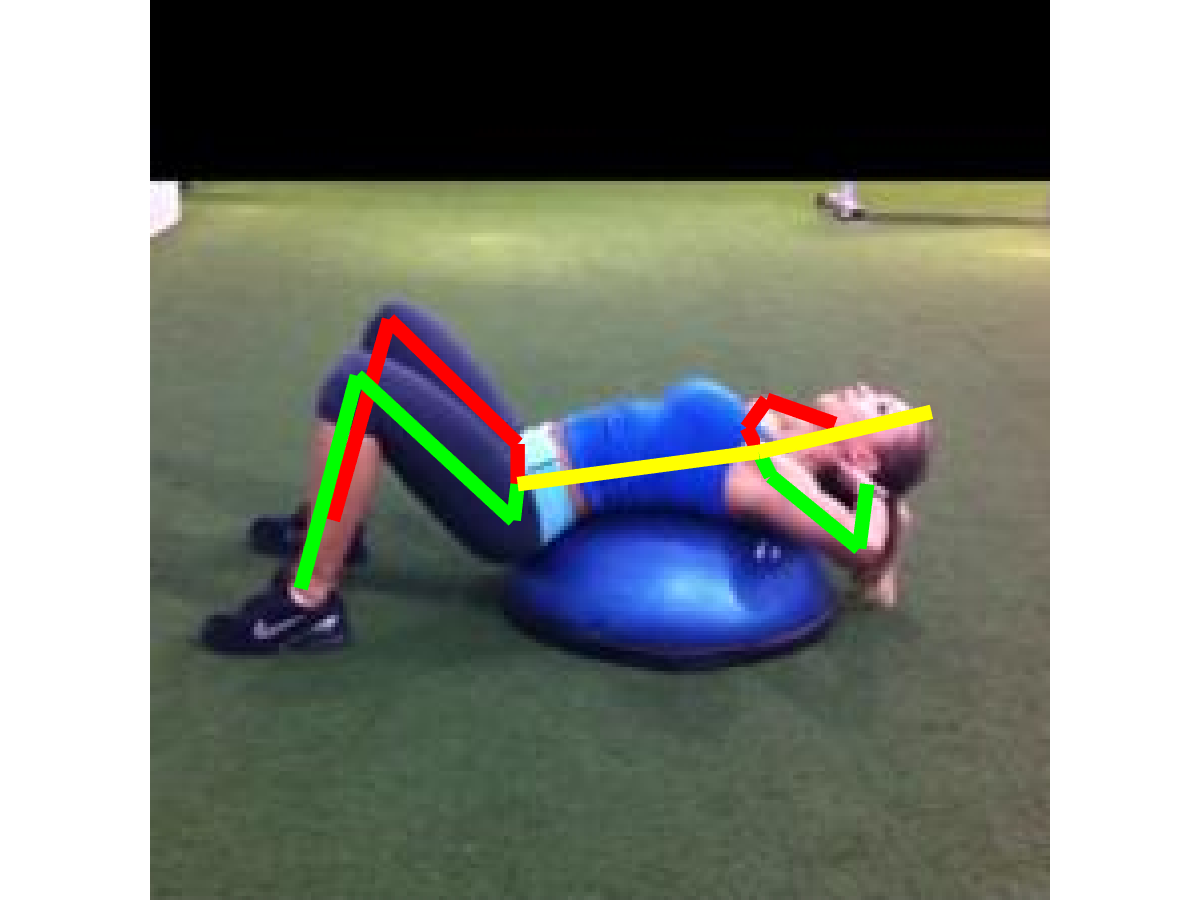} \\
\includegraphics[height = 0.1\textheight, width = 0.2\textwidth, keepaspectratio = true]{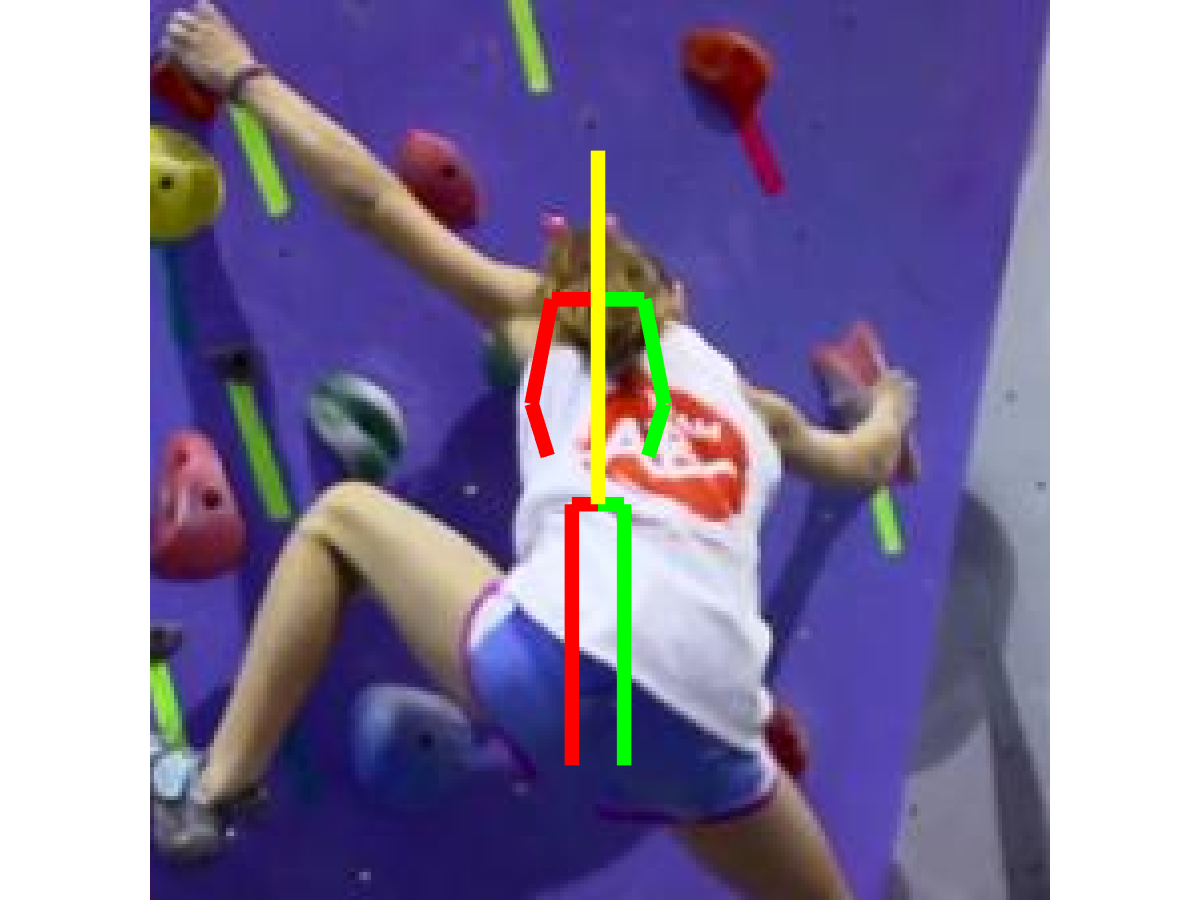} &
\includegraphics[height = 0.1\textheight, width = 0.2\textwidth, keepaspectratio = true]{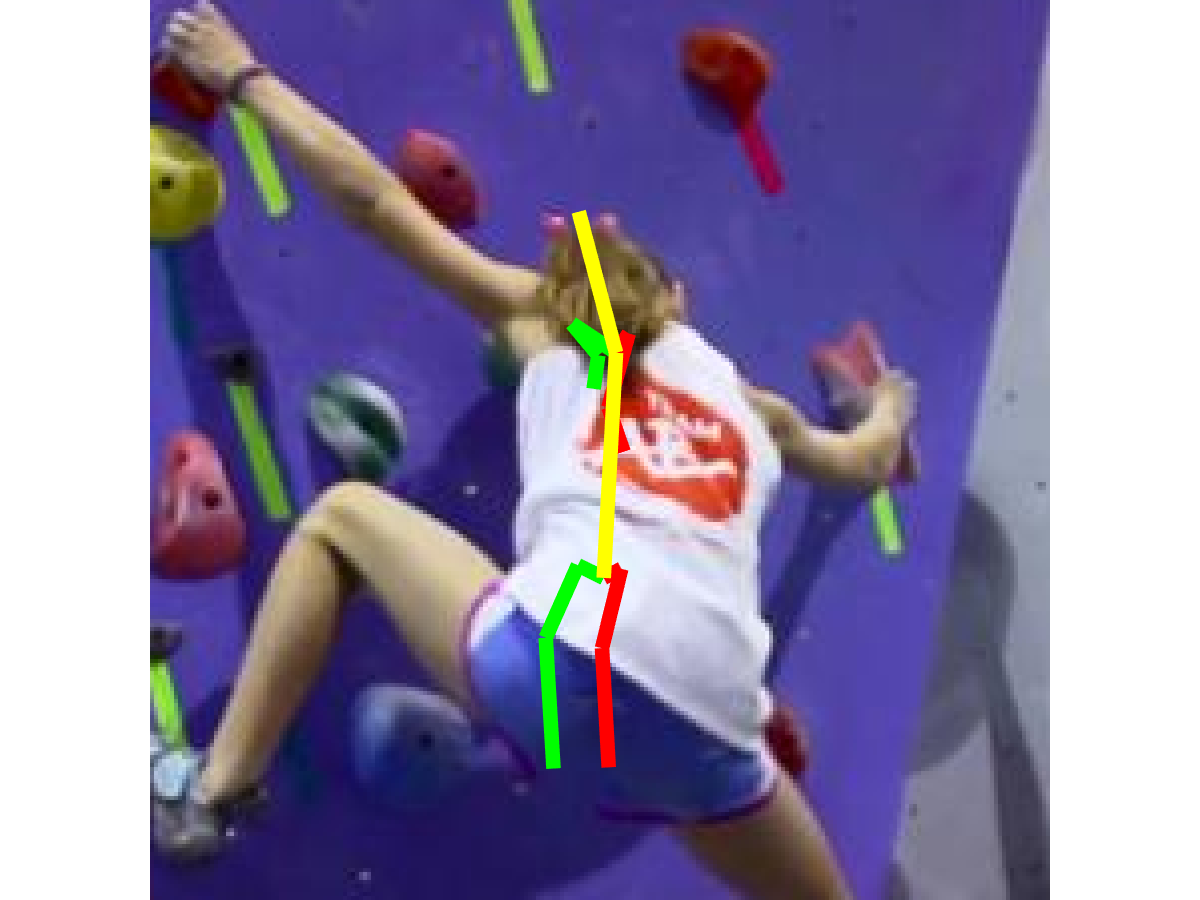} &
\includegraphics[height = 0.1\textheight, width = 0.2\textwidth, keepaspectratio = true]{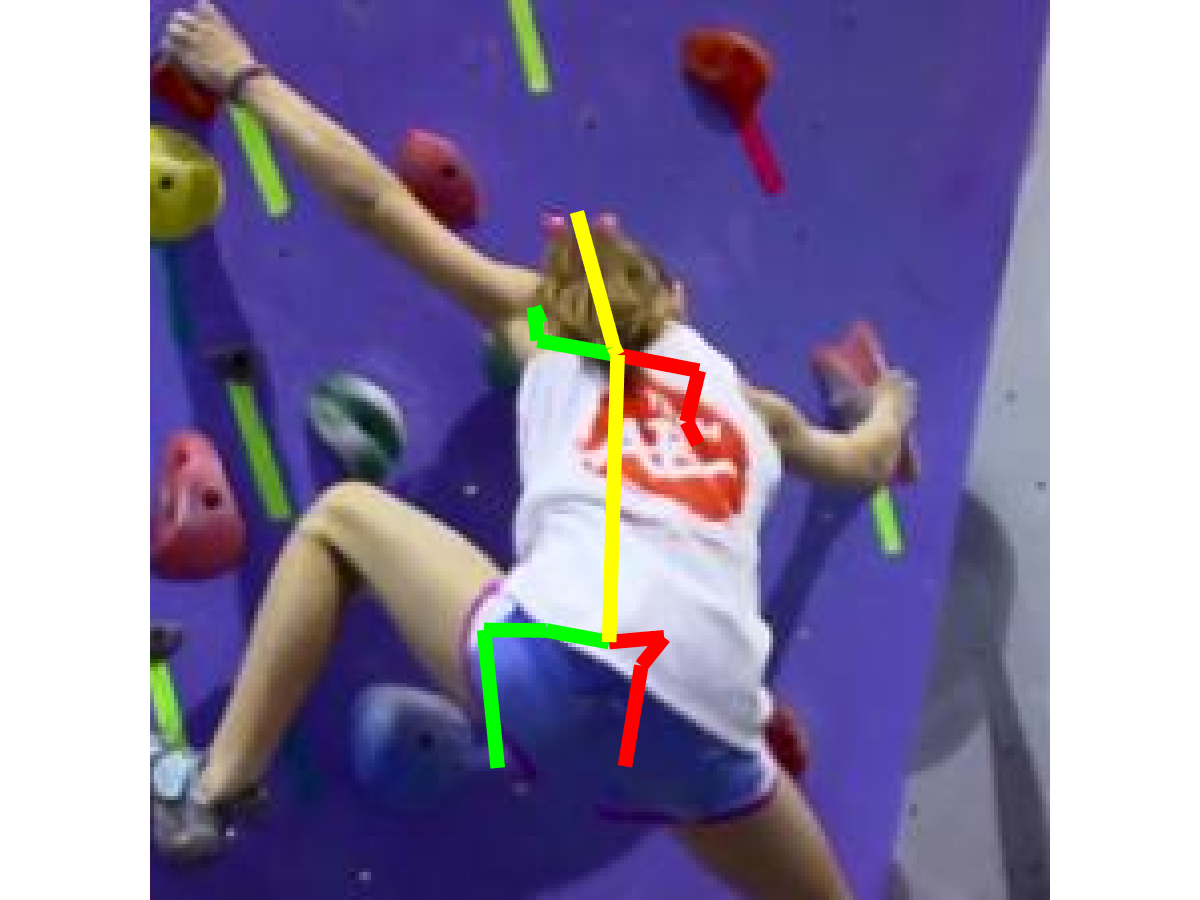} & 
\includegraphics[height = 0.1\textheight, width = 0.2\textwidth, keepaspectratio = true]{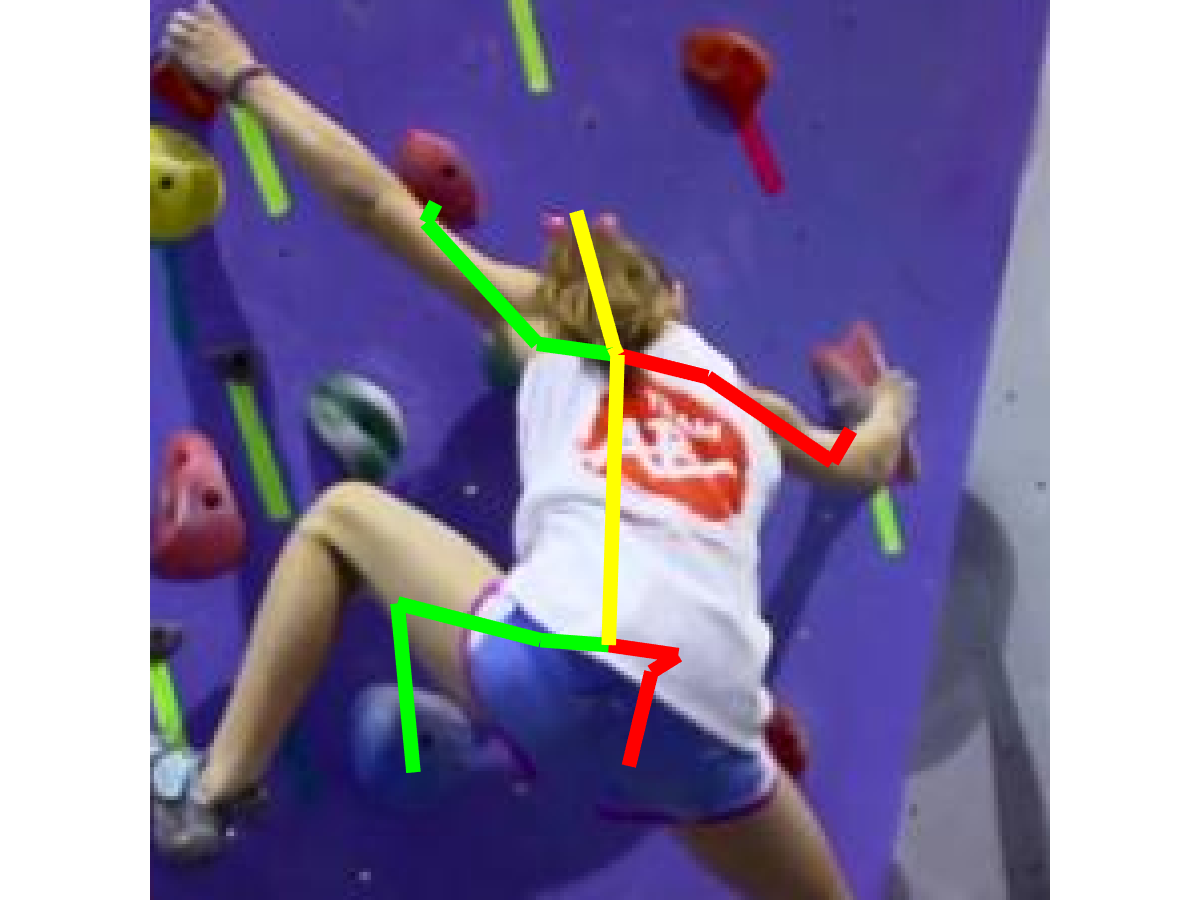} & 
\includegraphics[height = 0.1\textheight, width = 0.2\textwidth, keepaspectratio = true]{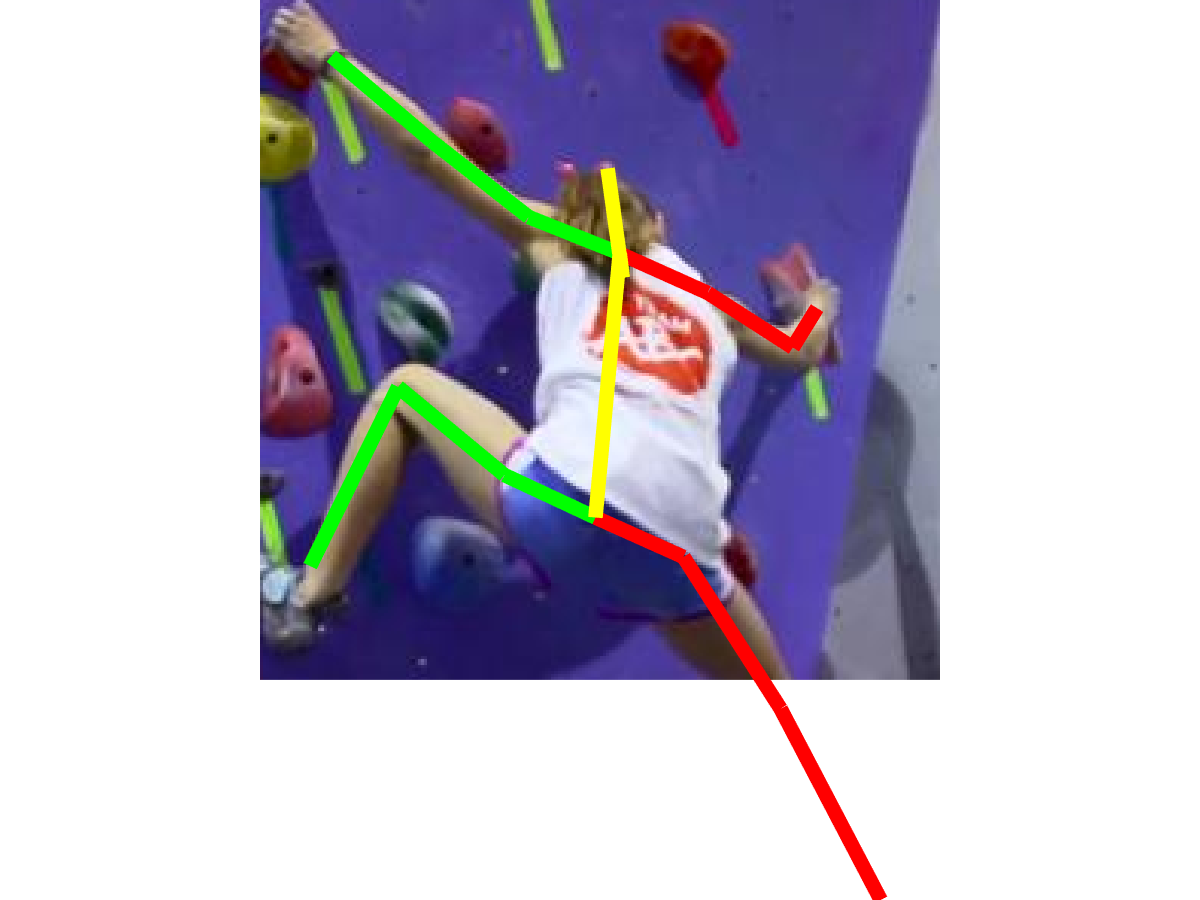} \\
\includegraphics[height = 0.1\textheight, width = 0.2\textwidth, keepaspectratio = true]{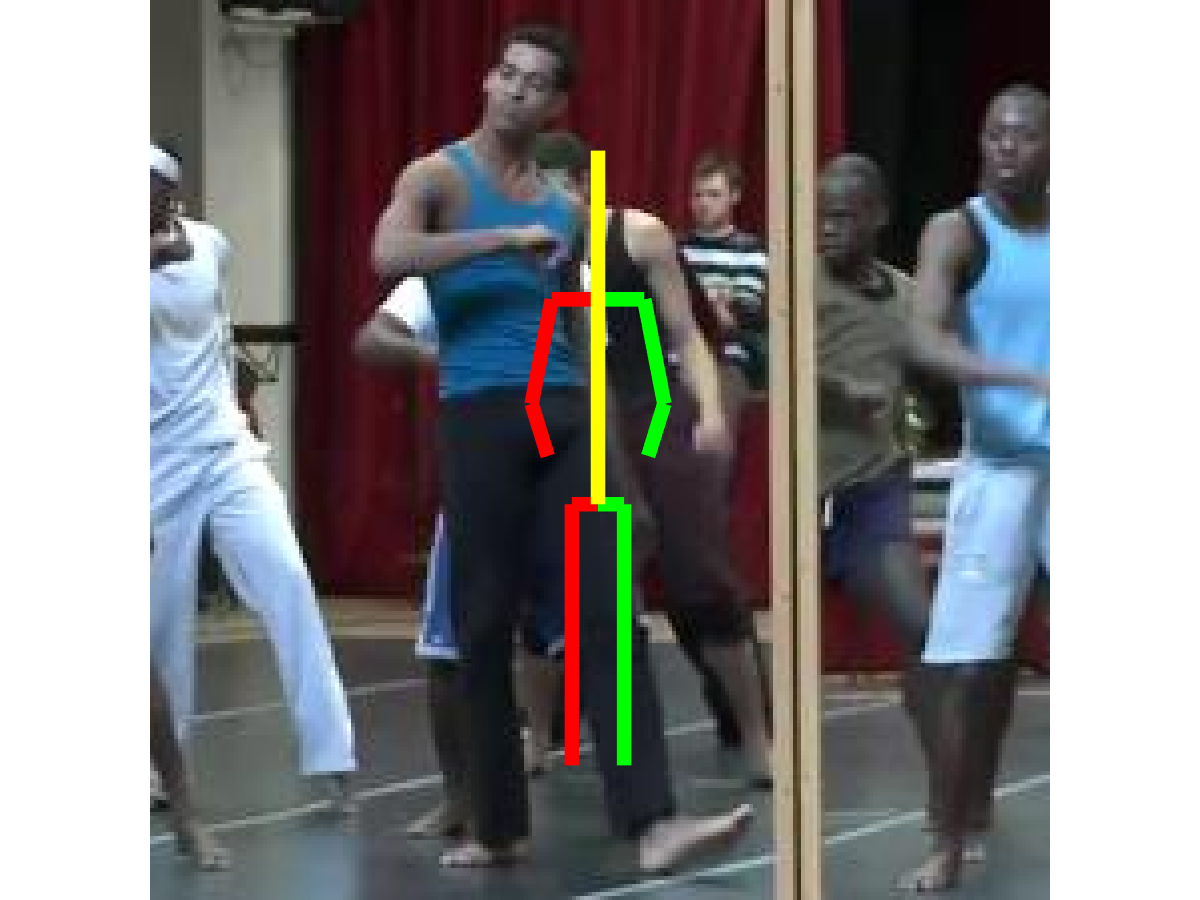} &
\includegraphics[height = 0.1\textheight, width = 0.2\textwidth, keepaspectratio = true]{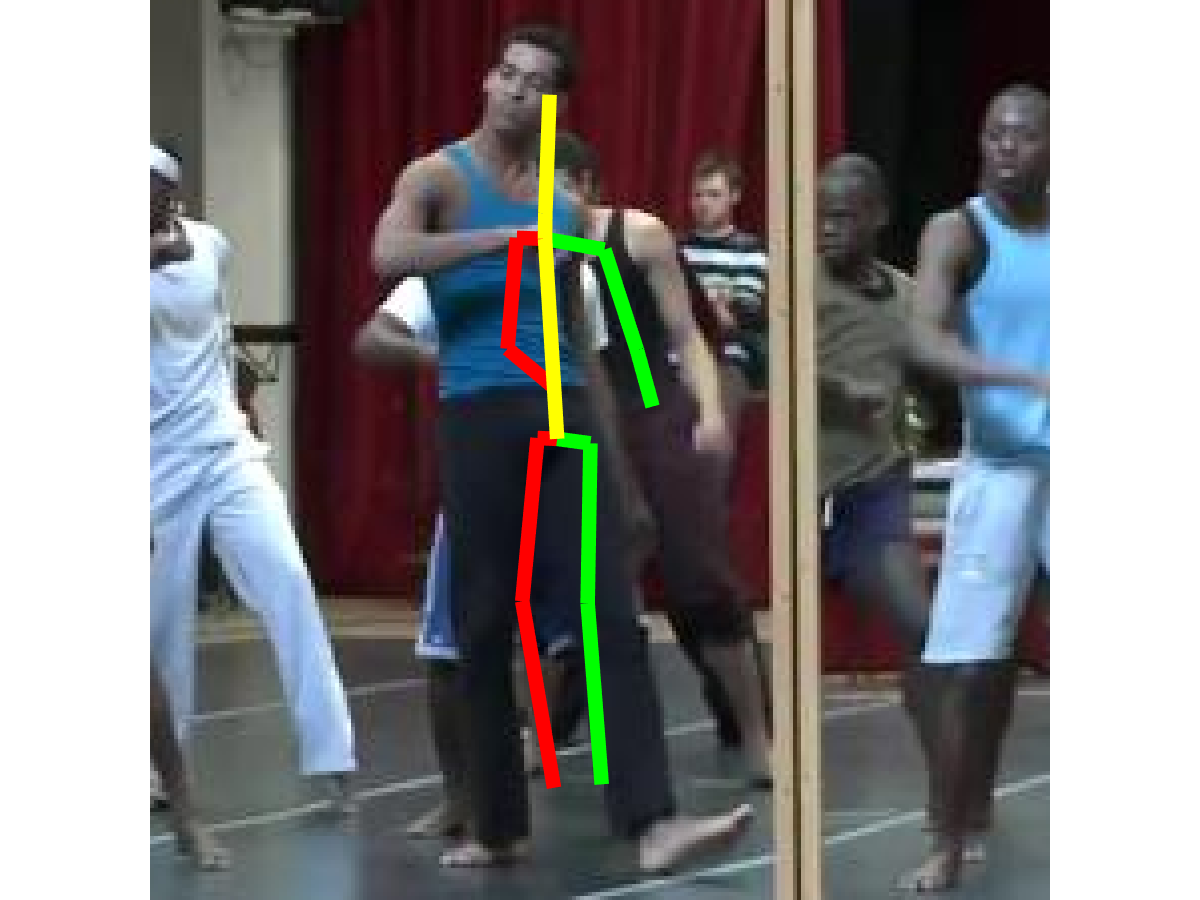} &
\includegraphics[height = 0.1\textheight, width = 0.2\textwidth, keepaspectratio = true]{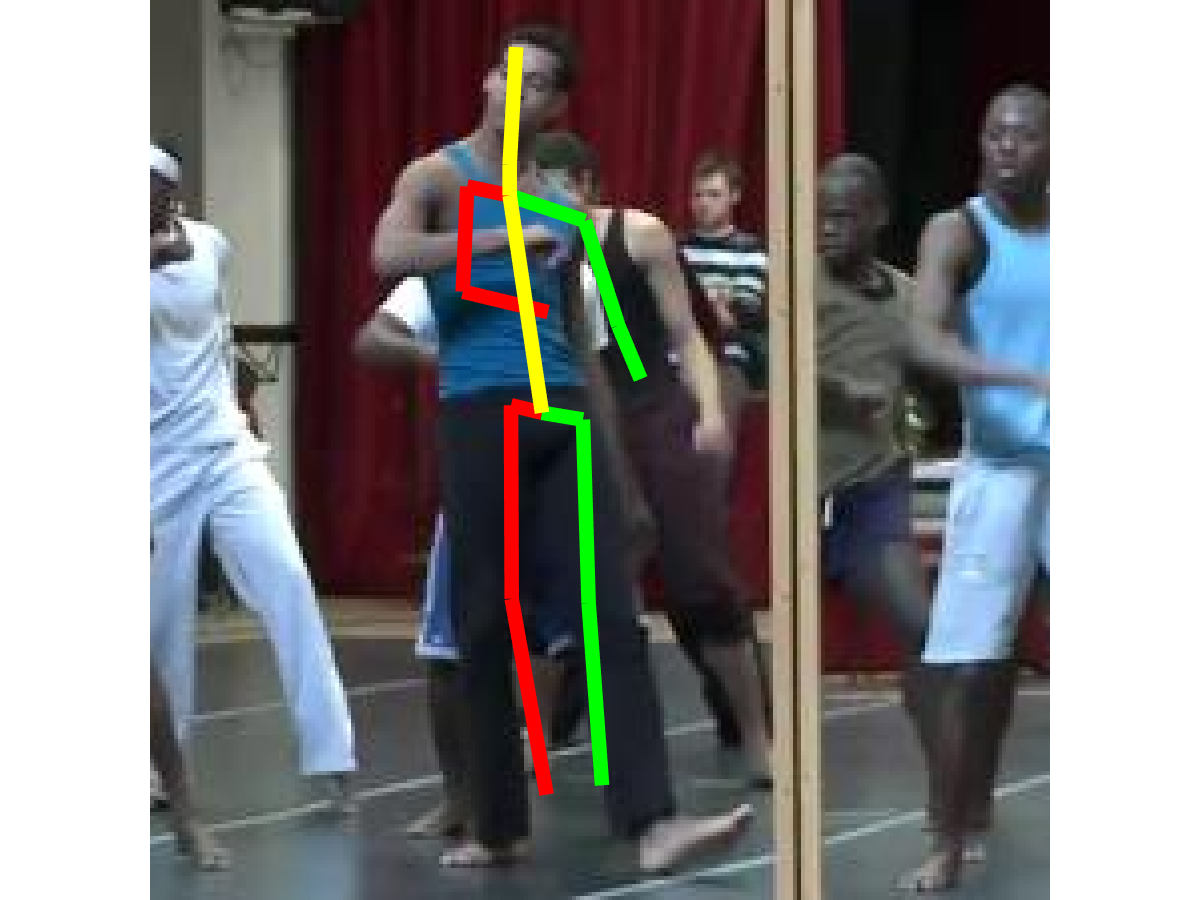} & 
\includegraphics[height = 0.1\textheight, width = 0.2\textwidth, keepaspectratio = true]{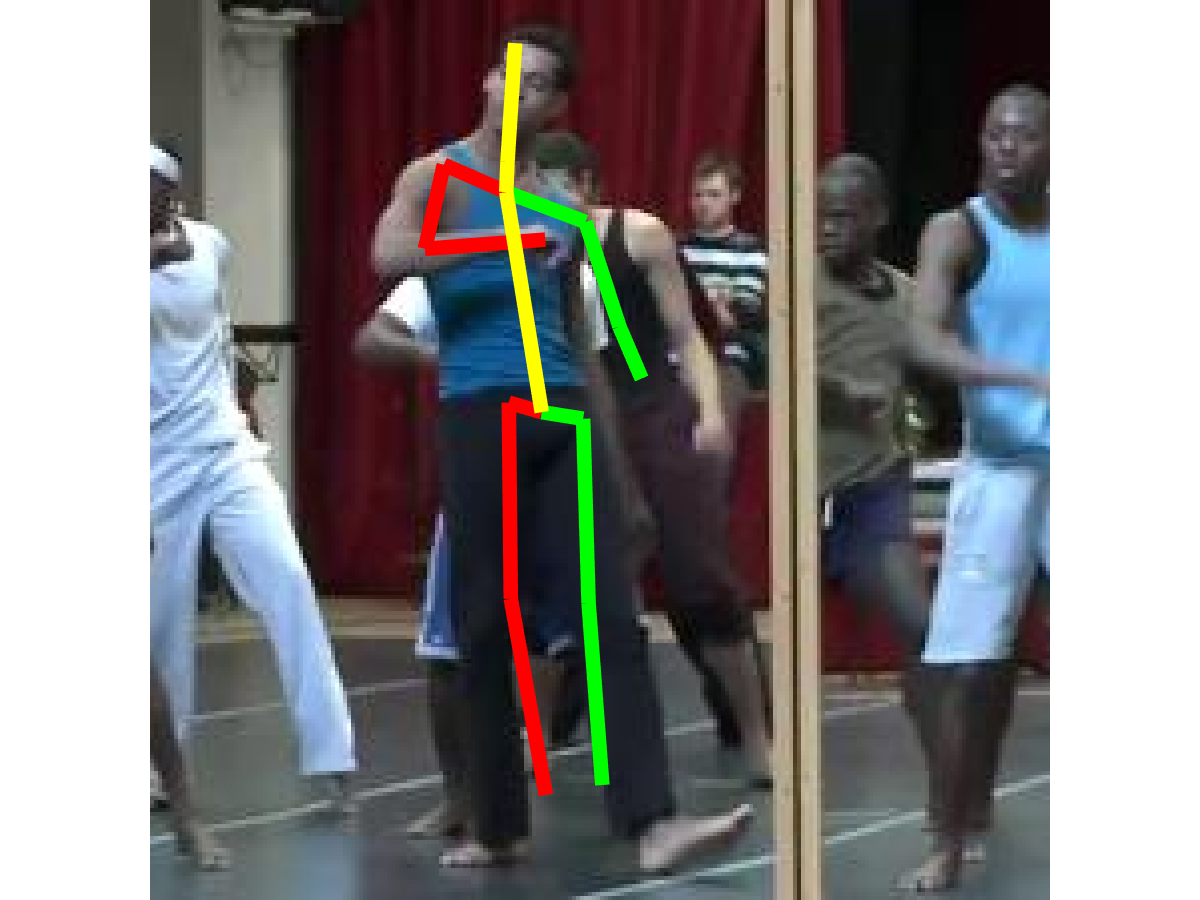} & 
\includegraphics[height = 0.1\textheight, width = 0.2\textwidth, keepaspectratio = true]{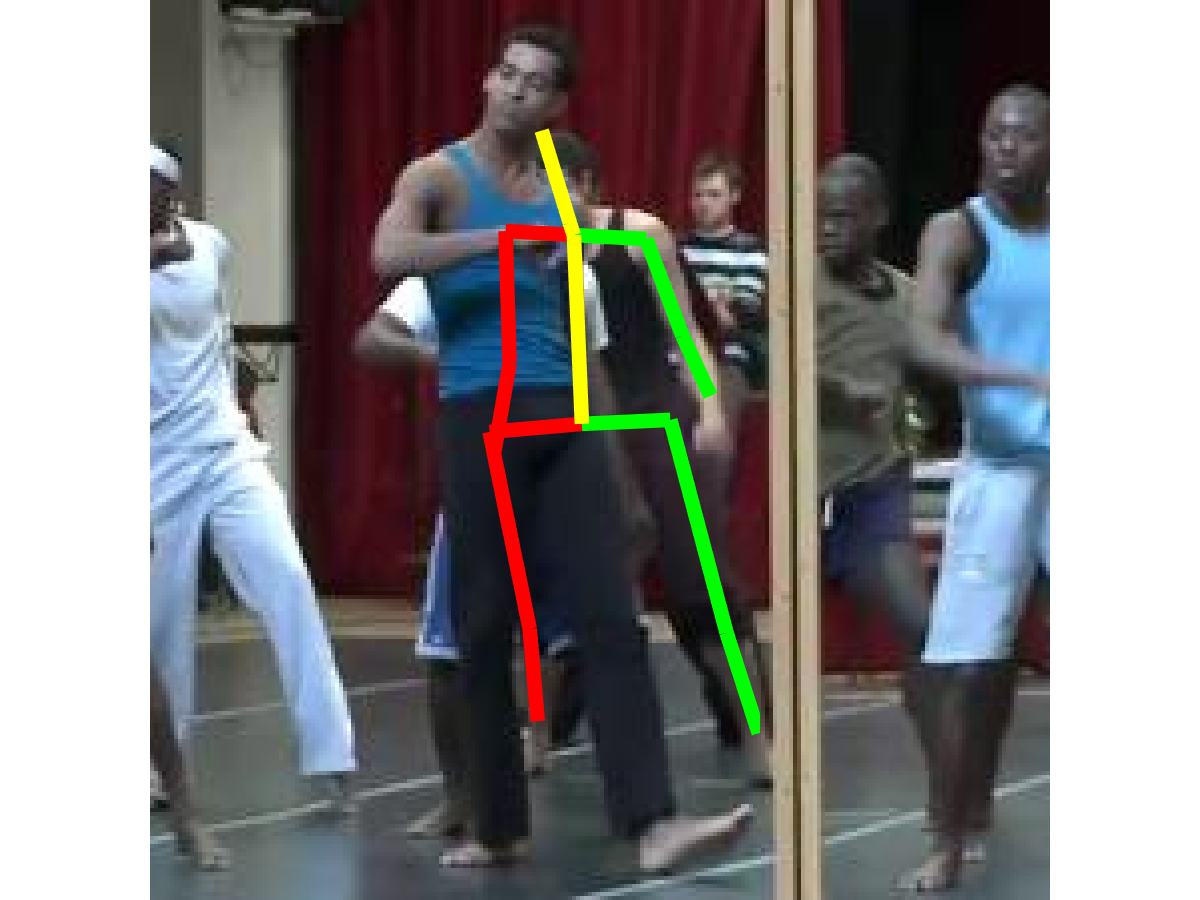} \\
\end{tabular}
  \caption{Example poses obtained using the proposed method IEF on the MPII validation set. From left to right we show the sequence of corrections the method makes -- on the right is the ground truth pose, including annotated occluded keypoints, which are not evaluated. Note that IEF is robust to left-right ambiguities and is able to rotate the initial pose by up to 180º (first and fifth row), can align across occlusions (second and third rows) and can handle scale variation (second, fourth and fifth rows) and truncation (fifth row). The bottom two rows show failure cases. In the first one, the predicted configuration captures the gist of the pose but is misaligned and not scaled properly. The second case shows several people closely interacting and the model aligns to the wrong person. The black borders show padding. Best seen in color and with zoom.}
\end{figure*}

\section{Conclusions}

While standard ConvNets offer hierarchical representations that can capture the patterns of images at multiple levels of abstraction, the outputs are typically modeled as flat image or pixel-level 1-of-K labels, or slightly more complicated hand-designed representations. We aimed in this paper to mitigate this asymmetry by introducing Iterative Error Feedback (IEF), which extends hierarchical representation learning to output spaces, while leveraging at heart the same machinery. IEF works by, in broad terms, moving the emphasis from the problem of \textit{predicting} the state of the external world to one of \textit{correcting}  the expectations about it, which is achieved by introducing a simple feedback connection in standard models.

In our pose estimation working example we opted for feeding pose information only into the first layer of the ConvNet for the sake of simplicity. This information may also be helpful for mid-level layers, so as to modulate not only edge detection, but also processes such as junction detection or contour completion which advanced feature extractors may need to compute. We also have only experimented so far feeding back "images" made up of Gaussian distributions. There may be more powerful ways to render top-down pose information  using parametrized computational blocks (e.g. deconvolution) that can then be learned jointly with the rest of the model parameters using standard backpropagation. This is desirable in order to attack problems with higher-dimensional output spaces such as 3D human pose estimation \cite{shakhnarovich2003fast,Sigal:IJCV:11} or segmentation.

\section*{Acknowledgement}{This work was supported in part by ONR MURI N00014-14-1-0671 and N00014-10-1-0933. Jo\~{a}o Carreira was partially supported by the Portuguese Science Foundation, FCT, under grant SFRH/BPD/84194/2012. Pulkit Agrawal was partially supported by a Fulbright Science and Technology  Fellowship. We gratefully acknowledge NVIDIA  corporation for the donation of
Tesla GPUs for this research. We thank Georgia Gkioxari and Carl Doersch for helpful comments.}

{\small
	\bibliographystyle{ieee}
	\bibliography{cvpr16_final_version.bib}

\begin{thebibliography}{10}\itemsep=-1pt

\bibitem{agrawal2014pixels}
P.~Agrawal, D.~Stansbury, J.~Malik, and J.~L. Gallant.
\newblock Pixels to voxels: Modeling visual representation in the human brain.
\newblock {\em arXiv preprint arXiv:1407.5104}, 2014.

\bibitem{andriluka20142d}
M.~Andriluka, L.~Pishchulin, P.~Gehler, and B.~Schiele.
\newblock 2d human pose estimation: New benchmark and state of the art
  analysis.
\newblock In {\em Computer Vision and Pattern Recognition (CVPR), 2014 IEEE
  Conference on}, pages 3686--3693. IEEE, 2014.

\bibitem{Chatfield14}
K.~Chatfield, K.~Simonyan, A.~Vedaldi, and A.~Zisserman.
\newblock Return of the devil in the details: Delving deep into convolutional
  nets.
\newblock In {\em British Machine Vision Conference}, 2014.

\bibitem{chen2014learning}
L.-C. Chen, A.~G. Schwing, A.~L. Yuille, and R.~Urtasun.
\newblock Learning deep structured models.
\newblock {\em arXiv preprint arXiv:1407.2538}, 2014.

\bibitem{chen2014articulated}
X.~Chen and A.~L. Yuille.
\newblock Articulated pose estimation by a graphical model with image dependent
  pairwise relations.
\newblock In {\em Advances in Neural Information Processing Systems}, pages
  1736--1744, 2014.

\bibitem{cootes1995active}
T.~F. Cootes, C.~J. Taylor, D.~H. Cooper, and J.~Graham.
\newblock Active shape models-their training and application.
\newblock {\em Computer vision and image understanding}, 61(1):38--59, 1995.

\bibitem{daume2009search}
H.~Daum{\'e}~III, J.~Langford, and D.~Marcu.
\newblock Search-based structured prediction.
\newblock {\em Machine learning}, 75(3):297--325, 2009.

\bibitem{DollarCVPR10pose}
P.~Doll\'ar, P.~Welinder, and P.~Perona.
\newblock Cascaded pose regression.
\newblock In {\em CVPR}, 2010.

\bibitem{fan2015combining}
X.~Fan, K.~Zheng, Y.~Lin, and S.~Wang.
\newblock Combining local appearance and holistic view: Dual-source deep neural
  networks for human pose estimation.
\newblock {\em arXiv preprint arXiv:1504.07159}, 2015.

\bibitem{Fan_2015_CVPR}
X.~Fan, K.~Zheng, Y.~Lin, and S.~Wang.
\newblock Combining local appearance and holistic view: Dual-source deep neural
  networks for human pose estimation.
\newblock June 2015.

\bibitem{citeulike:8460806}
D.~J. Felleman and D.~C. Van~Essen.
\newblock {Distributed Hierarchical Processing in the Primate Cerebral Cortex}.
\newblock {\em Cerebral Cortex}, 1(1):1--47, Jan. 1991.

\bibitem{fleuret-geman-rr2007}
F.~Fleuret and D.~Geman.
\newblock Stationary features and cat detection.
\newblock Technical Report Idiap-RR-56-2007, 0 2007.

\bibitem{fukushima1980neocognitron}
K.~Fukushima.
\newblock Neocognitron: A self-organizing neural network model for a mechanism
  of pattern recognition unaffected by shift in position.
\newblock {\em Biological cybernetics}, 36(4):193--202, 1980.

\bibitem{Gilbert2007677}
C.~D. Gilbert and M.~Sigman.
\newblock Brain states: Top-down influences in sensory processing.
\newblock {\em Neuron}, 54(5):677 -- 696, 2007.

\bibitem{hariharan2014hypercolumns}
B.~Hariharan, P.~Arbel{\'a}ez, R.~Girshick, and J.~Malik.
\newblock Hypercolumns for object segmentation and fine-grained localization.
\newblock {\em arXiv preprint arXiv:1411.5752}, 2014.

\bibitem{citeulike:506679}
J.~M. Hupe, A.~C. James, B.~R. Payne, S.~G. Lomber, P.~Girard, and J.~Bullier.
\newblock {Cortical feedback improves discrimination between figure and
  background by V1, V2 and V3 neurons}.
\newblock {\em Nature}, 394(6695):784--787, Aug. 1998.

\bibitem{ionescu2014iterated}
C.~Ionescu, J.~Carreira, and C.~Sminchisescu.
\newblock Iterated second-order label sensitive pooling for 3d human pose
  estimation.
\newblock In {\em Computer Vision and Pattern Recognition (CVPR), 2014 IEEE
  Conference on}, pages 1661--1668. IEEE, 2014.

\bibitem{jaderberg2014deep}
M.~Jaderberg, K.~Simonyan, A.~Vedaldi, and A.~Zisserman.
\newblock Deep structured output learning for unconstrained text recognition.
\newblock {\em ICLR 2015}, 2014.

\bibitem{Johnson10}
S.~Johnson and M.~Everingham.
\newblock Clustered pose and nonlinear appearance models for human pose
  estimation.
\newblock In {\em Proceedings of the British Machine Vision Conference}, 2010.
\newblock doi:10.5244/C.24.12.

\bibitem{kass1988snakes}
M.~Kass, A.~Witkin, and D.~Terzopoulos.
\newblock Snakes: Active contour models.
\newblock {\em International journal of computer vision}, 1(4):321--331, 1988.

\bibitem{ventral}
B.~C. U. L. M.~M. Kravitz~DJ, Saleem~KS.
\newblock The ventral visual pathway: An expanded neural framework for the
  processing of object quality.
\newblock volume 17(1), 2013.

\bibitem{LammeRoelfaema00}
V.~A.~F. Lamme and P.~R. Roelfaema.
\newblock the distinct modes of vision offered by feedforward and recurrent
  processing.
\newblock {\em Trends in Neurosciences}, 23:571, 2000.

\bibitem{lecun1998gradient}
Y.~LeCun, L.~Bottou, Y.~Bengio, and P.~Haffner.
\newblock Gradient-based learning applied to document recognition.
\newblock {\em Proceedings of the IEEE}, 86(11):2278--2324, 1998.

\bibitem{li2013fixed}
Q.~Li, J.~Wang, Z.~Tu, and D.~P. Wipf.
\newblock Fixed-point model for structured labeling.
\newblock In {\em Proceedings of the 30th International Conference on Machine
  Learning (ICML-13)}, pages 214--221, 2013.

\bibitem{NIPS2014_5542}
V.~Mnih, N.~Heess, A.~Graves, and K.~Kavukcuoglu.
\newblock Recurrent models of visual attention.
\newblock In Z.~Ghahramani, M.~Welling, C.~Cortes, N.~Lawrence, and
  K.~Weinberger, editors, {\em Advances in Neural Information Processing
  Systems 27}, pages 2204--2212. Curran Associates, Inc., 2014.

\bibitem{nowozin2011structured}
S.~Nowozin and C.~H. Lampert.
\newblock Structured learning and prediction in computer vision.
\newblock {\em Foundations and Trends{\textregistered} in Computer Graphics and
  Vision}, 6(3--4):185--365, 2011.

\bibitem{oberweger2015training}
M.~Oberweger, P.~Wohlhart, and V.~Lepetit.
\newblock Training a feedback loop for hand pose estimation.
\newblock In {\em Proceedings of the IEEE International Conference on Computer
  Vision}, pages 3316--3324, 2015.

\bibitem{pfister2015flowing}
T.~Pfister, J.~Charles, and A.~Zisserman.
\newblock Flowing convnets for human pose estimation in videos.
\newblock In {\em Proceedings of the IEEE International Conference on Computer
  Vision}, pages 1913--1921, 2015.

\bibitem{pishchulin2013poselet}
L.~Pishchulin, M.~Andriluka, P.~Gehler, and B.~Schiele.
\newblock Poselet conditioned pictorial structures.
\newblock In {\em Computer Vision and Pattern Recognition (CVPR), 2013 IEEE
  Conference on}, pages 588--595. IEEE, 2013.

\bibitem{pishchulin2013strong}
L.~Pishchulin, M.~Andriluka, P.~Gehler, and B.~Schiele.
\newblock Strong appearance and expressive spatial models for human pose
  estimation.
\newblock In {\em International Conference on Computer Vision (ICCV), 2013 IEEE
  Conference on}, pages 3487--3494. IEEE, 2013.

\bibitem{ramakrishna2014pose}
V.~Ramakrishna, D.~Munoz, M.~Hebert, J.~A. Bagnell, and Y.~Sheikh.
\newblock Pose machines: Articulated pose estimation via inference machines.
\newblock In {\em Computer Vision--ECCV 2014}, pages 33--47. Springer
  International Publishing, 2014.

\bibitem{shakhnarovich2003fast}
G.~Shakhnarovich, P.~Viola, and T.~Darrell.
\newblock Fast pose estimation with parameter-sensitive hashing.
\newblock In {\em Computer Vision, 2003. Proceedings. Ninth IEEE International
  Conference on}, pages 750--757. IEEE, 2003.

\bibitem{Sigal:IJCV:11}
L.~Sigal, M.~Isard, H.~Haussecker, and M.~J. Black.
\newblock Loose-limbed people: Estimating {3D} human pose and motion using
  non-parametric belief propagation.
\newblock {\em International Journal of Computer Vision}, 98(1):15--48, May
  2011.

\bibitem{simonyan2014very}
K.~Simonyan and A.~Zisserman.
\newblock Very deep convolutional networks for large-scale image recognition.
\newblock {\em arXiv preprint arXiv:1409.1556}, 2014.

\bibitem{stollenga2014deep}
M.~F. Stollenga, J.~Masci, F.~Gomez, and J.~Schmidhuber.
\newblock Deep networks with internal selective attention through feedback
  connections.
\newblock In {\em Advances in Neural Information Processing Systems}, pages
  3545--3553, 2014.

\bibitem{Szegedy_2015_CVPR}
C.~Szegedy, W.~Liu, Y.~Jia, P.~Sermanet, S.~Reed, D.~Anguelov, D.~Erhan,
  V.~Vanhoucke, and A.~Rabinovich.
\newblock Going deeper with convolutions.
\newblock June 2015.

\bibitem{Tompson_2015_CVPR}
J.~Tompson, R.~Goroshin, A.~Jain, Y.~LeCun, and C.~Bregler.
\newblock Efficient object localization using convolutional networks.
\newblock June 2015.

\bibitem{NIPS2014_5573}
J.~J. Tompson, A.~Jain, Y.~LeCun, and C.~Bregler.
\newblock Joint training of a convolutional network and a graphical model for
  human pose estimation.
\newblock In Z.~Ghahramani, M.~Welling, C.~Cortes, N.~Lawrence, and
  K.~Weinberger, editors, {\em Advances in Neural Information Processing
  Systems 27}, pages 1799--1807. 2014.

\bibitem{toshev2014deeppose}
A.~Toshev and C.~Szegedy.
\newblock Deeppose: Human pose estimation via deep neural networks.
\newblock In {\em Computer Vision and Pattern Recognition (CVPR), 2014 IEEE
  Conference on}, pages 1653--1660. IEEE, 2014.

\bibitem{tsochantaridis2004support}
I.~Tsochantaridis, T.~Hofmann, T.~Joachims, and Y.~Altun.
\newblock Support vector machine learning for interdependent and structured
  output spaces.
\newblock In {\em Proceedings of the twenty-first international conference on
  Machine learning}, page 104. ACM, 2004.

\bibitem{tu2008auto}
Z.~Tu.
\newblock Auto-context and its application to high-level vision tasks.
\newblock In {\em Computer Vision and Pattern Recognition, 2008. CVPR 2008.
  IEEE Conference on}, pages 1--8. IEEE, 2008.

\bibitem{tulving1990priming}
E.~Tulving and D.~L. Schacter.
\newblock Priming and human memory systems.
\newblock {\em Science}, 247(4940):301--306, 1990.

\bibitem{vedaldi2014matconvnet}
A.~Vedaldi and K.~Lenc.
\newblock Matconvnet-convolutional neural networks for matlab.
\newblock {\em arXiv preprint arXiv:1412.4564}, 2014.

\bibitem{weiss2012structured}
D.~Weiss, B.~Sapp, and B.~Taskar.
\newblock Structured prediction cascades.
\newblock {\em arXiv preprint arXiv:1208.3279}, 2012.

\bibitem{wolpert1992stacked}
D.~H. Wolpert.
\newblock Stacked generalization.
\newblock {\em Neural networks}, 5(2):241--259, 1992.

\bibitem{wyatte:the}
D.~Wyatte, T.~Curran, and R.~C. O'Reilly.
\newblock The limits of feedforward vision: Recurrent processing promotes
  robust object recognition when objects are degraded.
\newblock {\em J. Cognitive Neuroscience}, pages 2248--2261, 2012.

\bibitem{xiong2013supervised}
X.~Xiong and F.~De~la Torre.
\newblock Supervised descent method and its applications to face alignment.
\newblock In {\em CVPR}, 2013.

\bibitem{YangR_CVPR_2011}
Y.~Yang and D.~Ramanan.
\newblock Articulated pose estimation with flexible mixtures-of-parts.
\newblock In {\em CVPR}, 2011.

\end{thebibliography}
}

\end{document}